\documentclass[journal]{IEEEtran}

\usepackage{graphicx}
\usepackage{multirow}
\usepackage{amssymb}
\usepackage{epstopdf}
\usepackage{hyperref}
\ifCLASSINFOpdf
\else
\fi
\begin{document}
%
\title{NUC-Net: Non-uniform Cylindrical Partition Network for Efficient LiDAR Semantic Segmentation}
%
%
%
\IEEEpubidadjcol
\author{Xuzhi Wang, Wei Feng,~\IEEEmembership{Member,~IEEE},
        Lingdong Kong, Liang Wan,~\IEEEmembership{Member,~IEEE}
\thanks{Xuzhi Wang is with College of Computer and Information Engineering, Tianjin Normal University, Tianjin, 300380 China. e-mail: wangxuzhi@tjnu.edu.cn Xuzhi Wang is also with Tianjin Key Laboratory of Student Mental Health and Intelligence Assessment.}

\thanks{ Wei Feng and Liang Wan are with the College of Intelligence and Computing, Tianjin University, Tianjin, 300350 China. (e-mail: wangxuzhi@tju.edu.cn; lwan@tju.edu.cn; wfeng@tju.edu.cn) }

\thanks{Lingdong Kong is with the Department of Computer Science, National University of Singapore, Singapore, 117597, Singapore. (e-mail: lingdong@comp.nus.edu.sg)}

\thanks{ Corresponding author: Xuzhi Wang (email: wangxuzhi@tju.edu.cn).\newline \newline}}

%
%

\markboth{IEEE TRANSACTIONS ON CIRCUITS AND SYSTEMS FOR VIDEO TECHNOLOGY,~Vol., No.~}%
{Shell \MakeLowercase{\textit{et al.}}: Bare Demo of IEEEtran.cls for IEEE Journals}
%



\maketitle

\begin{abstract}
LiDAR semantic segmentation plays a vital role in autonomous driving. Existing voxel-based methods for LiDAR semantic segmentation apply uniform partition to the 3D LiDAR point cloud to form a structured representation based on cartesian/cylindrical coordinates.  Although these methods show impressive performance, the drawback of existing voxel-based methods remains in two aspects: (1) it requires a large enough input voxel resolution, which brings a large amount of computation cost and memory consumption. (2) it does not well handle the unbalanced point distribution of LiDAR point cloud. In this paper, we propose a non-uniform cylindrical partition network named NUC-Net to tackle the above challenges.  Specifically, we propose the Arithmetic Progression of Interval (API) method to non-uniformly partition the radial axis and generate the voxel representation which is representative and efficient. Moreover, we propose a non-uniform multi-scale aggregation method to improve contextual information. Our method achieves state-of-the-art performance on SemanticKITTI and nuScenes datasets with much faster speed and much less training time. And our method can be a general component for LiDAR semantic segmentation, which significantly improves both the accuracy and efficiency of the uniform counterpart by $4 \times$ training faster and  $2 \times$ GPU memory reduction and $3 \times$ inference speedup. We further provide theoretical analysis towards understanding why NUC is effective and how point distribution affects performance. Code is available at \href{https://github.com/alanWXZ/NUC-Net}{https://github.com/alanWXZ/NUC-Net}.
\end{abstract}


%
\IEEEpeerreviewmaketitle

\section{Introduction}

\IEEEPARstart{L}{iDAR} semantic segmentation aims to provide the per-point semantic information of the surrounding environment, which plays a significant role in autonomous driving. For real-world applications, the safety of passengers is crucial and the hardware resources are limited. Therefore, it is vital to design an accurate and efficient LiDAR semantic segmentation method.

Data representation matters for point cloud processing. Since LiDAR point clouds are large-scale and not in a regular format, most recent works ~\cite{LiDAR_Cylindrical,LiDAR_AF-S3Net,LiDAR_PolarNet} transform the irregular data into the voxel-based representation which balances the efficiency and effectiveness. The pioneer voxel-based methods ~\cite{point_VoxelNet,voxel_OctNet,voxel_VoxNet} subdivide the 3D space into equally spaced voxels based on cartesian coordinates and aggregate the information in each voxel. In this way, it avoids the time-consuming sampling and grouping operations of point-based methods ~\cite{point_pointnet,point_pointnet++} and obtains a structured data representation that can be processed by efficient sparse convolution ~\cite{point_submanifold}.
\begin{figure*}

	\begin{center}
		\centering
		\includegraphics[scale=0.41]{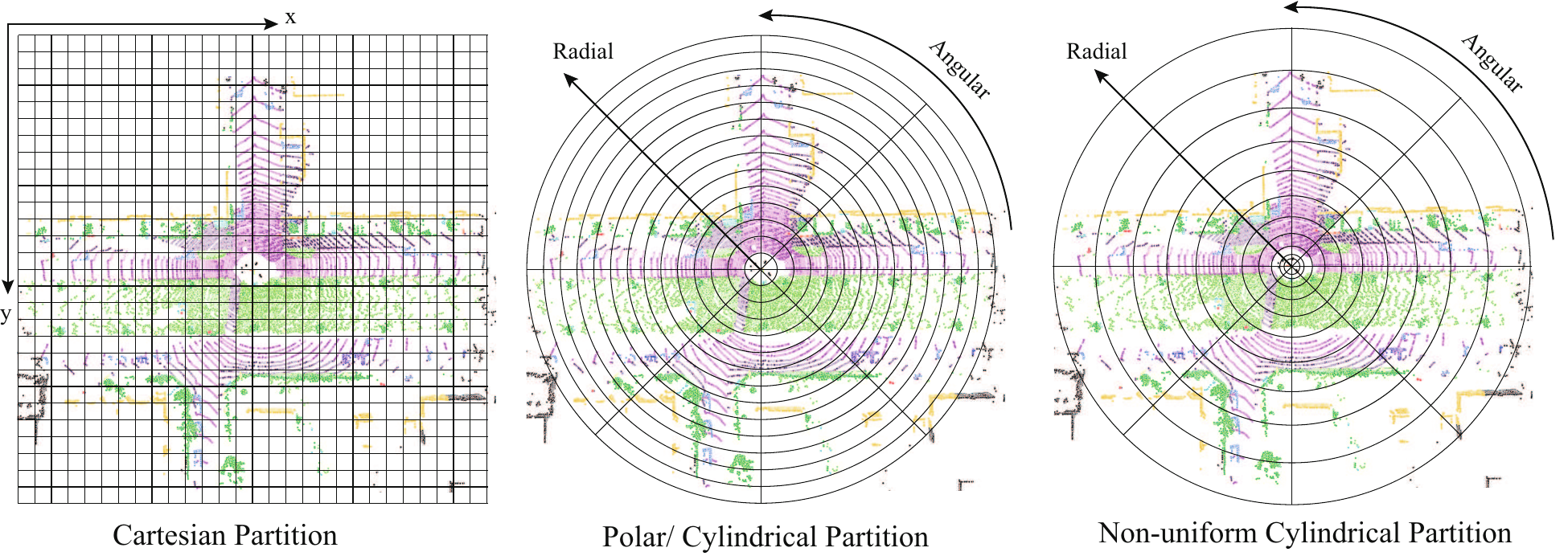}
	\end{center}
	\caption{Different partition mechanisms for voxel-based methods. From left to right: Cartesian Partition ~\protect\cite{LiDAR_NAS,LiDAR_2DPASS,point_VoxelNet,LiDAR_LinK}, Polar/Cylindrical Partition~\protect\cite{LiDAR_Cylindrical,LiDAR_PolarNet} and the proposed Non-uniform Cylindrical Partition(NUC).  The radial intervals of our method are increasing with the distance to the origin to balance the point distribution and reduce the computation cost. Note that we omit the $z$ axis for clarity and show the point clouds in a bird-eye-view.}
	
	\label{fig:non-uniform}

\end{figure*}

More recently, the voxel-based methods, PolarNet~\cite{LiDAR_PolarNet} and Cylinder3D~\cite{LiDAR_Cylindrical}, respect the fact that LiDAR point clouds have varied densities. To be specific, regions that are close to the LiDAR have much dense points and regions that are far away from the LiDAR have much sparse points. Therefore, they propose the Polar~\cite{LiDAR_PolarNet}/Cylindrical~\cite{LiDAR_Cylindrical} Partition method which forms a structured representation by uniformly dividing the 3D space based on polar/cylindrical coordinate as shown in Fig.~\ref{fig:non-uniform}. This representation achieves a more balanced point distribution compared with the cartesian-based method, which shows impressive performance for LiDAR semantic segmentation.

However, PolarNet~\cite{LiDAR_PolarNet} projects the 3D point cloud into the 2D bird's-eye view(BEV) to speed up LiDAR semantic segmentation, which loses the 3D geometry structure during projection. Cylinder3D~\cite{LiDAR_Cylindrical} achieves impressive performance by retaining the 3D geometric structure. However, the computation cost and GPU memory overhead limit the usage in real-world applications.

In an attempt to jointly tackle the above challenges, we propose a more representative and efficient voxel representation which is formed by non-uniformly partitioning the radial axis based on cylindrical coordinates. Intuitively, we allocate the nearby dense points with small radial intervals and allocate the faraway sparse points with large radial intervals as shown in Fig.~\ref{fig:non-uniform}.  Therefore, the fine details of the nearby points are retained and the more sparse faraway regions are represented by fewer voxels to save computation cost and GPU memory. Specifically, we design several versions of non-uniform partition mechanisms and adopt the proposed Arithmetic Progression of Interval (API) method which achieves the best performance. Moreover, we propose a non-uniform multi-scale aggregation method to encode the multi-scale features.

\IEEEpubid{\begin{minipage}{\textwidth}\ \centering
		Copyright \copyright 20xx IEEE. Personal use of this material is permitted. \\
		However, permission to use this material for any other purposes must be obtained 
		from the IEEE by sending an email to pubs-permissions@ieee.org.
\end{minipage}}

\IEEEpubidadjcol
Our method is substantially novel compared with existing works. (1) our method forms the voxel representation by non-uniformly partitioning the 3D space along the radial axis. This is different from the OctNet~\cite{voxel_OctNet} and SphereFormer~\cite{LiDAR_Spherical}. OctNet uniformly partitions all non-empty regions using octrees, while our method adopts varied partition intervals for different regions. SphereFormer adopts the uniform partition to form the voxel representation and proposes the exponential splitting(a non-uniform partition mechanism) for position encoding of the transformer. Note that SphereFormer does not bring speed improvement. By non-uniformly partitioning the radial axis, our method greatly reduces the computation cost and achieves a more balanced point distribution to boost both efficiency and performance. (2) We propose Arithmetic Progression of Interval and non-uniform multi-scale aggregation, wihch is a novel design. (3) We further provide theoretical analysis towards understanding why the proposed non-uniform cylindrical partition is effective and how point distribution affects performance from the perspective of CNN.

Different from existing efficient LiDAR semantic segmentation methods using range/BEV maps ~\cite{LiDAR_Rangenet++,LiDAR_PolarNet,LiDAR_RPVNet}, knowledge distillation ~\cite{LiDAR_PVKD} and lightweight architecture ~\cite{LiDAR_DRINet,LiDAR_RandLA-Net}, the proposed NUC-Net speeds up by the more efficient and representative NUC representation, which shows superior performance and complementary to existing efficient methods.

The contributions of our work are summarised as follows:

(1) We propose the non-uniform cylindrical representation which is both representative and efficient. We design several non-uniform partition mechanisms and adopt the proposed Arithmetic Progression of Interval (API) method. Further, we propose the non-uniform multi-scale feature aggregation method to boost contextual information.

(2) Our method achieves state-of-the-art performance on SemanticKITTI and nuScenes datasets with much faster speed and much less training time. And our method can be a general component for LiDAR semantic segmentation, which could improve both the accuracy and efficiency of the uniform counterparts by $4 \times$ training faster and  $2 \times$ GPU memory reduction and $3 \times$ inference speedup.  These improvements stem from the NUC partitioning strategy, which enhances the accuracy while reducing the input resolution by a factor of four.

(3) We are the first to present a thorough analysis of how different voxel partition mechanisms affect performance. We provide both a theoretical and experimental analysis of our method. We show that the superior performance is attributed to the NUC-Net which reduces the encoding error in the nearby regions and enlarges the receptive field in the faraway regions.

\section{Related Works}

\subsection{LiDAR Semantic Segmentation}

Data representation is vital for point cloud processing~\cite{t_lidar_compression,t_lidar_fusion,t_lidar_seg,t_video_lidar, ffnet}. According to different data representations, existing works can be categorized into the following categories:

\textbf{Point-based Methods} The pioneering work of the point-based method is PointNet~\cite{point_pointnet} which directly processes point cloud based on an MLP-based network. Although point-based methods have shown effectiveness to process small-scale point clouds, they are inefficient to process large-scale point clouds. To cope with this problem, RandLA-Net~\cite{LiDAR_RandLA-Net} proposes an effective LiDAR segmentation method based on the effective local feature aggregation module and random sampling strategy.

\textbf{Range-based Methods}
The range-based methods project the 3D point cloud into the spherical projection. Range-based representation can be processed by standard 2D convolution which is very efficient. SqueezeSeg ~\cite{LiDAR_SqueezeSegV3} projects 3D points into spherical projection and boosts the semantic segmentation network by the SAC module.  RangeNet++~\cite{LiDAR_Rangenet++} proposes a novel post-processing algorithm to deal with discretization errors and blurry CNN outputs. RangeViT~\cite{LiDAR_RangeViT} exploits the knowledge from 2D images to boost range-based LiDAR segmentation using the vision transformer.

\begin{figure*}
	\begin{center}
		\centering
		
		\includegraphics[scale=0.4]{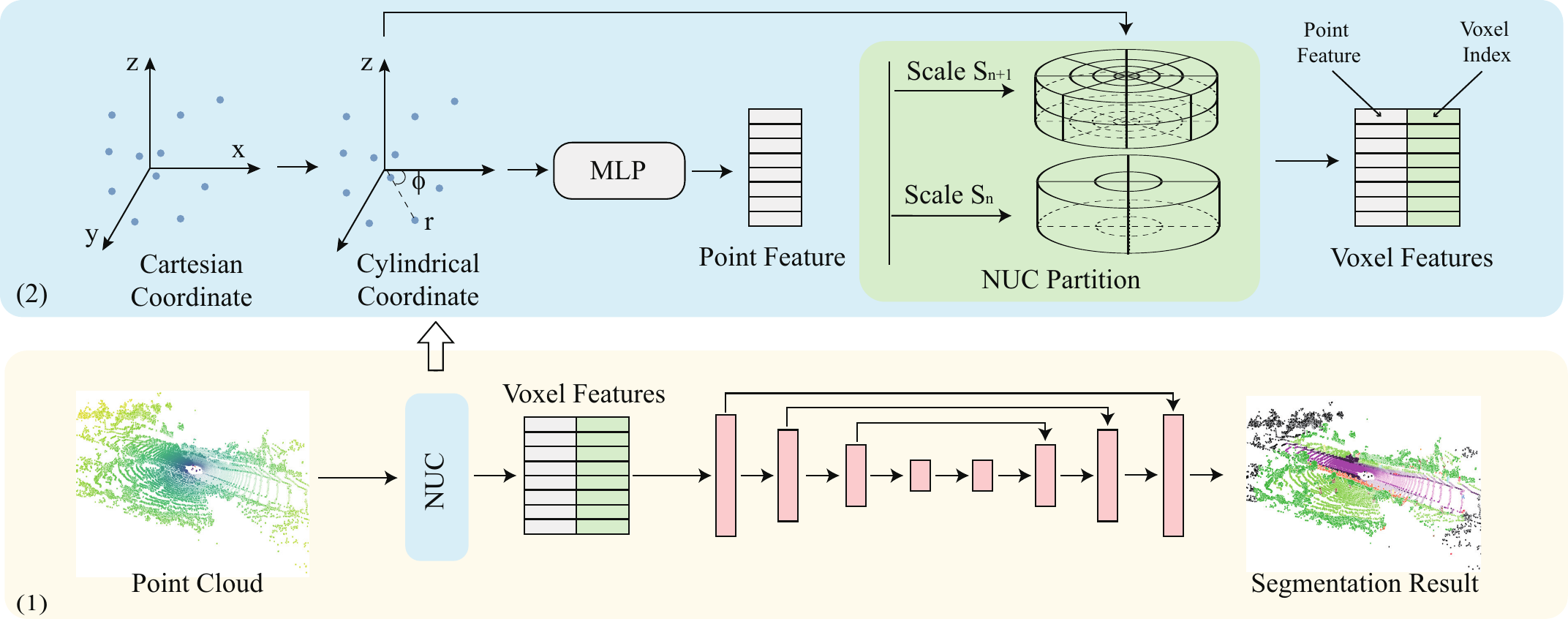}
	\end{center}
	\caption{(1) The framework of our method. We first obtain the voxel-wise feature using the proposed non-uniform cylindrical partition method. Then, the voxel-vise feature is fed into a sparse 3D encoder-decoder convolution network to predict the results.   (2) The detail of NUC Representation. The point cloud is first converted to the cylindrical coordinate and processed by a set of MLP layers to obtain the point-wise features. The point-wise features are assigned to the corresponding voxels based on the non-uniform cylindrical partition method to form the voxel-wise feature.}

	\label{fig:framework}
	\vspace{-8pt} 
\end{figure*}

\textbf{Voxel-based Methods} 
Voxel-based methods convert unregular points into structured 3D voxel/grids representation to use 3D/2D convolution.
Early voxel-based methods use BEV representation with cartesian coordinates to convert points into voxels and apply 2D convolution for semantic segmentation. Considering the varied density of LiDAR point cloud, PolarNet~\cite{LiDAR_PolarNet} designs a partition method based on polar BEV coordinates. Cylinder3D ~\cite{LiDAR_Cylindrical} proposes a 3D cylindrical partition network to solve the problem of losing 3D details of BEV-based methods. AF2S3Net~\cite{LiDAR_AF-S3Net} proposes a multi-branch attentive fusion module and an adaptive feature selection module to learn both global contexts and local details. LinK ~\cite{LiDAR_LinK} proposes a linear kernel method to achieve a wider-range perceptive field for LiDAR segmentation. SphereFormer~\cite{LiDAR_Spherical} designs radial window self-attention based on spherical coordinates and proposes exponential splitting for position encoding of transformer.

Existing voxel-based LiDAR segmentation methods apply uniform partition to form a voxel representation. Different from existing works, our method proposes the non-uniform partition to better fit the point distribution and speedup. The partition interval is varied according to the point density. Note that SphereFormer~\cite{LiDAR_Spherical} adopts uniform partition to form a voxel representation and embeds the proposed exponential splitting(a non-uniform mechanism) position encoding to each voxel. Our method can boost both efficiency and performance, while SphereFormer~\cite{LiDAR_Spherical} does not bring any speed improvement.

\textbf{Fusion-based Methods} Different representations have their own pros and cons, fusion-based methods aim at fusing different representations, such as point-based, projection-based and voxel-based, to explore a hybrid representation.  SPVNAS ~\cite{LiDAR_NAS} proposes a neural architecture search method to search for an optimal architecture for point-voxel LiDAR segmentation. DRINet~\cite{LiDAR_DRINet} proposes dual-representation learning with great flexibility at feature transferring and less computation cost. RPVNet~\cite{LiDAR_RPVNet} proposes a range-point-voxel fusion method aiming to synergize all three view's representations.

\subsection{Non-uniform Partition}

Non-uniform sampling strategies have been widely explored in the following aspects. 1) Non-uniformly sampling the regular kernels/grids ~\cite{sampling_AdaInt}, such as deformable convolution ~\cite{sampling_deformable_kernels} and adaptive downsampling methods~\cite{sampling_PointRend}. 2) Sampling the non-uniform distributed data, such as performing farthest point sampling (FPS) on a point cloud~\cite{point_pointnet}.

The most related works to our approach are OctNet~\cite{voxel_OctNet} and SphereFormer~\cite{LiDAR_Spherical}. To exploit the sparsity property, OctNet~\cite{voxel_OctNet} uses the octree partition which recursively subdivides the 3D space into octants. For each iteration, OctNet uniformly partitions the octree nodes that contain a data point in its domain. OctNet does not consider the varied point distribution of point cloud. SphereFormer~\cite{LiDAR_Spherical} adopts the uniform partition for point cloud voxelization and proposes the exponential splitting for position encoding of the transformer. Since SphereFormer only adds the exponential splitting position encoding for each uniform partitioned voxel, it does not bring speed improvement.

Different from existing methods, our approach focuses on non-uniformly partitioning the 3D point cloud and aggregates the information in each cell to transform the irregular point cloud into regular voxels. The non-uniform partition mechanism better fits the point distribution and reduces the input voxel resolution, which could boost both efficiency and performance.

\section{Method}

\subsection{Overview}

The framework of our method is shown in Fig.~\ref{fig:framework}. The raw point clouds are first processed by a set of MLP layers. The features extracted from MLP are processed by the Arithmetic Progression of Interval (API) method to generate a representative and computationally efficient voxel representation. Then, the voxel features are processed by the designed 3D sparse convolution network to obtain the segmentation results. We show the details of our method in the following.

\subsection{Uniform Partition}

We first revisit the two uniform partition methods ~\cite{LiDAR_Cylindrical,LiDAR_PolarNet} which show impressive performance based on cylindrical coordinates. Then, we formulate the volume of each cell of the cylindrical partition, which is core to handle the varied densities of the LiDAR point cloud. At last, we analyze the limitations of these methods based on the formulation.

\textbf{Uniform Polar/Cylindrical Partition} Polar and Cylindrical Partition methods take into account the imbalanced spatial distribution of LiDAR point cloud where nearby regions have much greater point density than far-away regions. The basic idea of Polar/Cylindrical Partition is to utilize a larger grid size to cover the far-away region and achieves a more balanced point distribution. Given the LiDAR point cloud, Polar/Cylinder-based methods first use a set of MLP layers to extract features and the point features in the same partitioned voxel are aggregated using the max pooling operation to form a fixed-length 3D voxel representation which can be process by the 3D network.

\textbf{Formulation} For polar/cylindrical partition, Cylinder3D~\cite{LiDAR_Cylindrical} first transforms the points $(x,y,z)$ based on cartesian coordinate to $(r,\phi,z)$ based on cylindrical coordinate. Suppose that the point cloud encompasses a 3D space with range $H,W,D$ and the input voxel resolution is $n_{r}*n_{\phi}*n_z$, where $n_{r}$ is the number of partitions along radial coordinate, $n_{\phi}$ is the number of partitions along angular coordinate and $n_{z}$ is the number of partitions along $z$ coordinate. 

As existing cylindrical partition method subdivides the point cloud into a set of equal intervals along each coordinate. The voxel size along radial coordinate, angular coordinate and z coordinate can be defined as $a=\sqrt{H^2+W^2}/n_r$, $b=2\pi/n_{\phi}$ and $h=D/n_{z}$, respectively. The volume of the $i,j,k$-th cell with size $a \times b \times h$ denoted as $V_{i,j,k}$ can be formulated as:

\begin{eqnarray}   
V_{ i,j,k}&=&\frac{h  \pi(((i+1)a)^2-(ia)^2)}{2 \pi /b}  \nonumber    \\
~&=&\frac{a^{2} b h (2i +1)}{2} ,
\end{eqnarray}
where $i \in [0, 1, ..., n_r], j \in [0, 1, ..., n_{\phi}], k\in [0, 1, ..., n_z]$ are the voxel index along $r$, $\phi$ and $z$ coordinates. We can observe that the volume $V_{i,j,k}$ is proportional to the index $i$ along radial direction using uniform cylindrical partition method, denoted as $V_{i,j,k} \propto i$.

\textbf{Limitation} As the intervals $a$, $b$ and $h$ are constant values predefined by voxel resolution $a \times b \times h$, the variation of $V_{i,j,k}$ only depends on $i$ ($V_{i,j,k} \propto i$). The rate of volume change defined by the uniform cylindrical partition differs significantly from the rate of point cloud density variation as shown in Fig.~\ref{fig:non-uniform} (b). Although the uniform cylindrical partition achieves impressive performance gain by achieving a more balanced point cloud distribution~\cite{LiDAR_PolarNet,LiDAR_Cylindrical}, it has two intrinsic limitations: (1) The imbalance of point distribution is quite large even though using the cylindrical partition method. (2) Adopting the same radial interval for sparse regions as in dense regions will lead to much computation cost.

\subsection{Non-Uniform Cylindrical Partition}

To address the aforementioned challenges, we resolve to further adjust the radial intervals to increase the voxel volume's growth rate along the radial direction, thereby adapting to the variation in the distribution of the point cloud. We propose the Arithmetic Progression of Interval (API) partition method by exploring various non-uniform partitioning mechanisms. For API method, the difference between the consecutive intervals of the radical axis is a constant value and the interval of the radial axis is increasing with the distance to the origin. This representation can achieve a more balanced point distribution as shown in Fig.~\ref{fig:non-uniform} (b) and save much computation cost. Based on API, the i-th interval along the radial direction can be defined as follows:
\begin{equation}
a^i=a^0+i \times d,
\end{equation}
where $a^0$ represents the first interval of radial axis, $a^{i}$ represents the (i+1)-th interval of radial axis and $d$ represents the common difference of successive members.

The volume of $V_{i,j,k}$ using the proposed method can formulated as:
\begin{small}
	\begin{eqnarray}    
	V_{i,j,k}&=&\frac{h  \pi(((i+1)(2a^0+id)/2)^2-(i(2a^0+(i-1)d)/2)^2)}{2 \pi /b}  \nonumber    \\
	~&\!\!\!\!\!\approx\!\!\!\!\!&\frac{ b h  d^2 i^3}{2},
	\end{eqnarray}
\end{small}
where $i$ is the voxel index along the radial direction, $a^0$ is the first interval of the radial axis, $d$ is the common difference of successive members. $V_{i,j,k}$ can be approximately obtained as we set $a^0$ to be a small positive value. The volume $V_{i,j,k}$ is proportional to $i^3$ denoted as $V_{i,j,k} \propto i^3$ using the proposed Arithmetic Progression of Interval method. Moreover, the API mechanism is more flexible compared with uniform partition methods by adjusting $a^0$ and $d$.

\begin{figure}
	\centering
	\includegraphics[width=7cm]{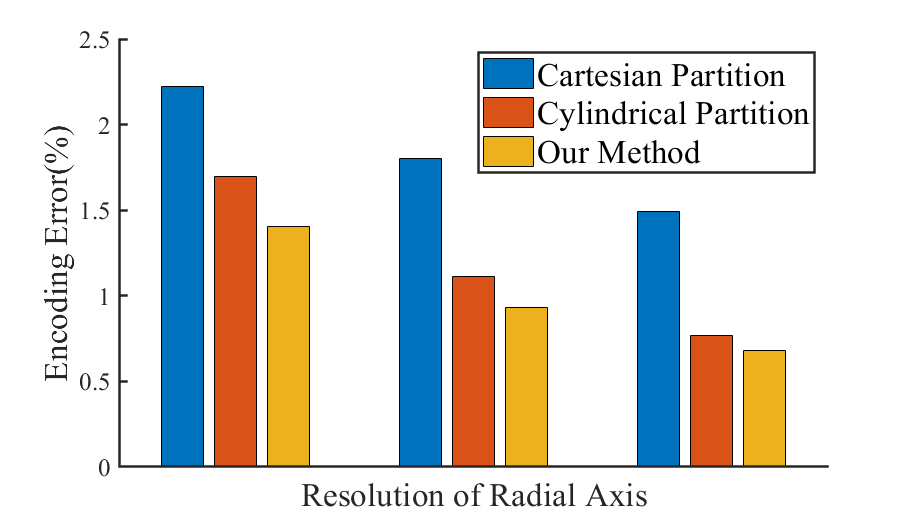} 
	\caption{Label encoding error of different partition methods. From left to right, the resolution of radial direction is 120, 240 and 480.}
	\label{fig:encoding_error}
	\vspace{-4pt} 
\end{figure}

As shown in Fig.~\ref{fig:non-uniform}~(b), our method achieves a more balanced point distribution compared with the uniform partition methods based on cylindrical coordinates and cartesian coordinates. Additionally, different voxel partitioning mechanisms can affect encoding errors which will compromise subsequent learning. When there are points of different categories within the same voxel, encoding errors occur. As shown in Fig.~\ref{fig:encoding_error}, our method achieves the lowest encoding errors across various voxel resolution settings.

\begin{figure}
	\centering
	\includegraphics[width=7cm]{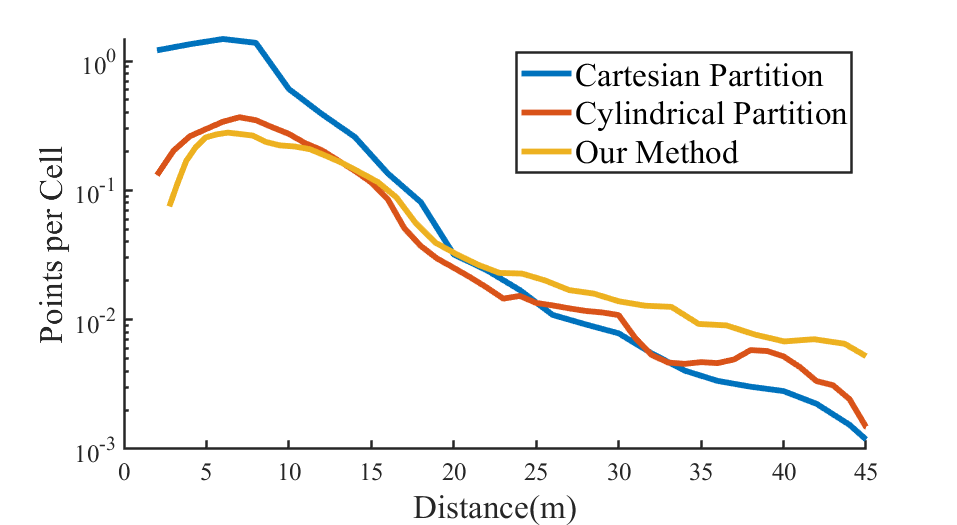} 
	\caption{The variation of points per cell with distance. Our method achieves a more balanced point distribution by allocating more voxels to the nearby area and less voxels to the far away area.  }
	\label{fig:point_density}
	\vspace{-0.5cm} 
\end{figure}

\subsection{Non-uniform Multi-scale Aggregation}
	\begin{figure}
	\centering
	\includegraphics[width=8cm]{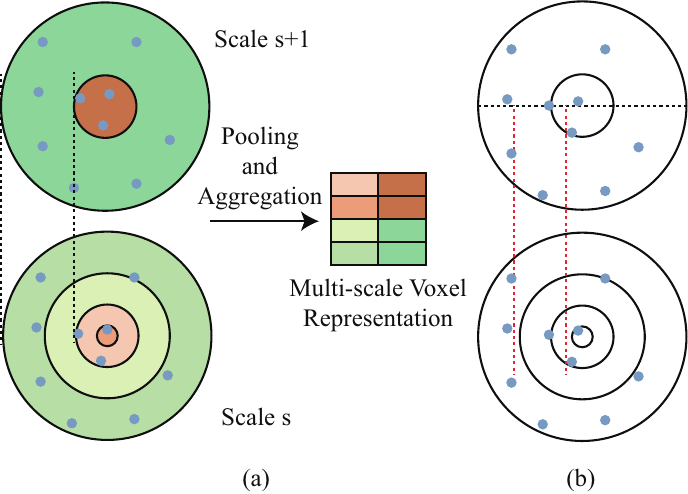} 
	\caption{(a) The Non-uniform Multi-scale Aggregation. (b) The inconsistencies caused by directly halving. The red line represents directly halving the interval at scale s+1 to obtain scale s, which causes inconsistencies between the intervals corresponding to scale s+1 and scale s, indicating that the red line does not align with the black circle.	}
	\label{fig:multi-scale}
	\vspace{-0.5cm}
\end{figure}
We further propose a non-uniform multi-scale aggregation method. Existing works~\cite{point_pointnet++,detection_HVNet} indicate that multi-scale information, offering context information with different receptive fields at various scales, is essential for segmentation task. However, due to the non-uniform radial intervals, directly applying existing multi-scale methods~\cite{detection_HVNet,LiDAR_DRINet} (downsampling for each voxel) would lead to mismatched feature ranges at different scales as shown in Fig.~\ref{fig:multi-scale}. For angular and z directions, we follow the existing multi-scale paradigm. For the interval of the radial axis, we formulate the interval of each scale as follows:
\begin{equation}
	a^{i}_{s}=2^{s}a^{0}+(2^{2s}i+2^{2s-1}-2^{s-1})d. 
\end{equation}
where $s \in S$ represents the scale s, $S$ represents the scale set.

After partitioning the space according to the non-uniform multi-scale partition, the point features belonging to the same voxel in each scale are transferred to the voxel feature by max pooling. Then, the voxel features of different scales are concatenated and fed into the 3D network.
\begin{equation}
V^{\psi} = \mathop{\mathrm{Concat}} \limits_{s = 1, ... ,t} \{ \mathop{\mathrm{Max}} \limits_{i=1,...,n} \{ \mathrm{h}(p_i^{s}) \} \},
\end{equation}
where $p_i^{s}$ represents the point set belonging to $V^{\psi}$, $V^{\psi}$ represents the concatenated multi-scale feature of the $\psi$ voxel, $\mathrm{Concat}(\cdot)$ represents tensor concatenation, $\mathrm{Max}(\cdot)$ represents max pooling and $h$ are multi-layer perceptron networks.

\subsection{Network Architecture}

\begin{figure}
	\centering
	\includegraphics[width=7cm]{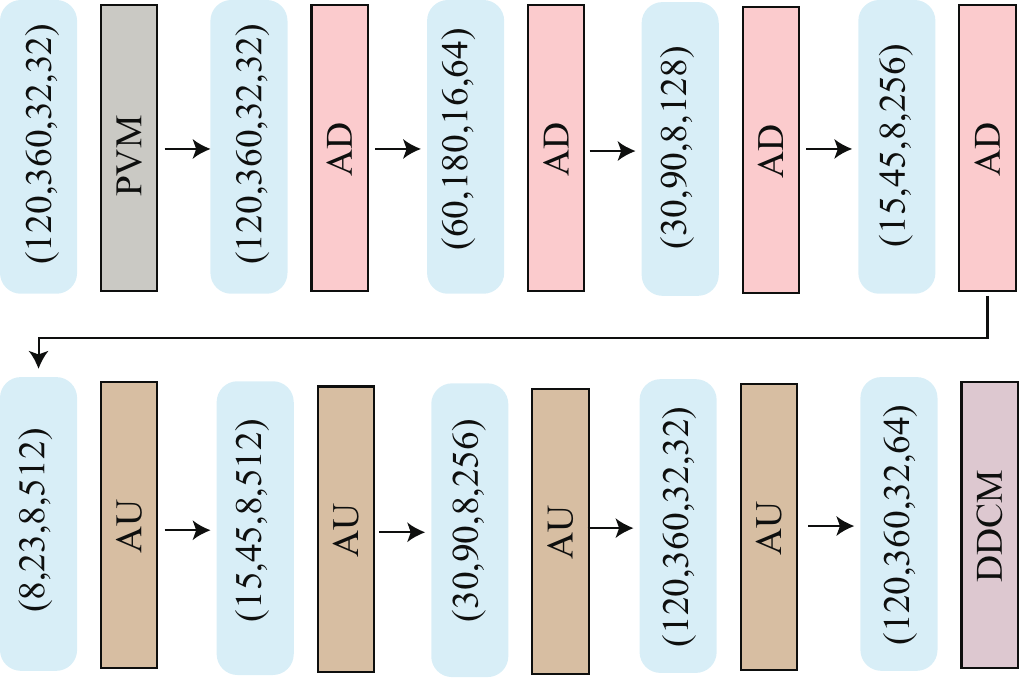} 
	\caption{The details of 3D Network architecture. `PVM' represents the Point-Voxel Module~\cite{LiDAR_PVCNN}. `AU' represents the Asymmetrical Upsample Block~\cite{LiDAR_Cylindrical}. `AD' represents the Asymmetrical Downsample Block~\cite{LiDAR_Cylindrical}. `DDCM' represents the Dimension-Deocoposition Context Module~\cite{LiDAR_Cylindrical}. The parameter of feature map are shown as (radial dimension, angular dimension, height dimension, channel dimension).}
	\label{fig:3d_architecture}
	\vspace{-0.5cm}
\end{figure}

After the above steps, the raw point cloud data are encoded into a fixed-length 3D voxel representation. The 3D voxel representations are processed by a set of sparse convolution layers~\cite{point_submanifold}. Specifically, as shown in Fig.~\ref{fig:3d_architecture}, we design 3D networks by using 4 encoding layers and 4 decoding layers as the 3D backbone network which is a modified voxel branch of Cylinder3D~\cite{LiDAR_Cylindrical}. We add the point-based branch~\cite{LiDAR_NAS} to the first layer of the 3D network, which could provide the per-point geometry information. We apply instance augmentation~\cite{aug_pointaugment,aug_panoptic-polarnet,LiDAR_RPVNet} to augment the less frequent objects in the scene, which can boost the performance and keep the runtime efficiency. The ground truth label of each voxel is encoded by the majority class label within the voxel.

\subsection{Loss Function}
We use the weighted cross-entropy loss and lovasz-softmax ~\cite{loss_lovasz-softmax} loss to maximize the point accuracy and the intersection-over-union score following the existing works ~\cite{LiDAR_Cylindrical,LiDAR_RandLA-Net}. The loss function can be represented as follows:

\begin{equation}
L=L_{lovasz}+L_{wce},
\end{equation}
where $L_{lovasz}$ represents the lovasz-softmax loss ~\cite{loss_lovasz-softmax} and $L_{wce}$ represents the weighted cross entropy loss. The weighted cross-entropy loss takes the class frequency into account to cope with the imbalanced class problem.
\section{Experiments}

\subsection{Datasets}
We evaluate the proposed method on SemanticKITTI~\cite{dataset_semanticKITTI} and nuScenes~\cite{dataset_nuScenes} datasets. 

\textbf{SemanticKITTI} SemanticKITTI~\cite{dataset_semanticKITTI} is a large-scale dataset for driving-scene. It consists of 22 point cloud sequences, where sequences 00 to 10 are the training set (sequence 08 is used as the validation set) and sequences 11 to 21 are the test set.  There are 23201 LiDAR point clouds for training and 20351 for testing. All sequences of the SemanticKITTI are derived from KITTI Vision Odeometry Benchmark and dense point-wise annotations are provided for LiDAR segmentation by SemanticKITTI.

\textbf{NuScenes}
NuScenes~\cite{dataset_nuScenes} is a large-scale dataset from cities in Boston and Singapore. Among the 1000 scenes, 700 of them are used for training, 150 of them are used for validation and 150 of them are used for testing. There are 16 classes utilized for semantic segmentation after merging similar classes.

To evaluate our method on SemanticKITTI and nuScenes dataset, we use mean intersection-over-union (mIoU) over all classes following the official guidance ~\cite{dataset_nuScenes,dataset_semanticKITTI}.

\subsection{Experiment Setups}

We use PyTorch to implement our experiments with GeForce RTX 3090. We design a non-uniform cylindrical partition network by the proposed Arithmetic Progression of Interval method (API). The first term $a^0$ is set as 0.05 and tolerance $d$ is set as 0.0062. The aggregation scale is set as 2. The model is trained with batch size 8 and the number of training epochs 40. During training, we use random scaling, flipping, rotation, translation and instance augmentation for data augmentation. Test-time ensemble~\cite{LiDAR_2DPASS} is conducted during the inference following~\cite{LiDAR_2DPASS,LiDAR_Cylindrical,LiDAR_NAS}. We set the input voxel resolution as $120\times 360 \times 32$, where the three dimensions represent the radius, angle and height, respectively. We train our network with Adam optimizer with a learning rate 0.001.

\subsection{Experimental Results}
\begin{table*}
	\setlength\tabcolsep{3pt}
	\tiny
	
	\begin{center}
		\resizebox{\linewidth}{!}{
			\begin{tabular}{c|c|c|c|ccccccccccccccccccc|c}
				\hline
				
				Methods&\rotatebox{90}{Input}&\rotatebox{90}{Resolution}&\rotatebox{90}{Mean IoU}&\rotatebox{90}{Car}& \rotatebox{90}{Bicycle}&\rotatebox{90}{Motorcycle}&\rotatebox{90}{Truck}&\rotatebox{90}{Other-vehicle}&\rotatebox{90}{Person}&\rotatebox{90}{Bicyclist}&\rotatebox{90}{Motorcyclist}&\rotatebox{90}{Road}&\rotatebox{90}{Parking}&\rotatebox{90}{Sidewalk}&\rotatebox{90}{Other-ground}&\rotatebox{90}{Building}&\rotatebox{90}{Fence}&\rotatebox{90}{Vegetation}&\rotatebox{90}{Trunk}&\rotatebox{90}{Terrain}&\rotatebox{90}{Pole}&\rotatebox{90}{traffic sign}&\rotatebox{90}{speed (ms)}\\
				\hline

				SqueezeSegV3
				~\cite{LiDAR_SqueezeSegV3}&L &\scalebox{0.8}{[64,2048]}&55.9&92.5&38.7&36.5&29.6&33.0&45.6&46.2&20.1&91.7&63.4&74.8&26.4&89.0&59.4&82.0&58.7&65.4&49.6&58.9&238\\

				RangeNet++
				~\cite{LiDAR_Rangenet++}&L &\scalebox{0.8}{[64,2048]}&52.2&91.4&25.7&34.4&25.7&23.0&38.3&38.8&4.8&91.8&65.0&75.2&27.8&87.4&58.6&80.5&55.1&64.6&47.9&55.9&83.3\\

				PointNet++
				~\cite{point_pointnet++}&L &-&20.1&53.9&1.9&0.2&0.9&0.2&0.9&1.0&0.0&72.0&18.7&41.8&5.6&62.3&16.9&46.5&13.8&30.0&6.0&8.9&5900\\

				RandLA-Net
				~\cite{LiDAR_RandLA-Net}&L &-&53.9&94.2&26.0&25.8&40.1&38.9&49.2&48.2&7.2&90.7&60.3&73.7&20.4&86.9&56.3&81.4&61.3&66.8&49.2&47.7&880\\

				PolarNet
				~\cite{LiDAR_PolarNet}&L &\scalebox{0.8}{[480,360,32]}&54.3&93.8&40.3&30.1&22.9&28.5&43.2&40.2&5.6&90.8&61.7&74.4&21.7&90.0&61.3&84.0&65.5&67.8&51.8&57.5& 62\\

				SPVNAS
				~\cite{LiDAR_NAS}&L &\scalebox{0.8}{[3200,3200,600]}&67.0&97.2&50.6&50.4&56.6&58.0&67.4&67.1&50.3&90.2&67.6&75.4&21.8&91.6&66.9&86.1&73.4&71.0&64.3&67.3&259\\
				
				Cylinder3D~\cite{LiDAR_Cylindrical}&L &\scalebox{0.8}{[480,360,32]}&67.8&97.1&67.6&64.0&59.0&58.6&73.9&67.9&36.0&91.4&65.1&75.5&32.3&91.0&66.5&85.4&71.8&68.5&62.6&65.6&171\\

				DRINet~\cite{LiDAR_RPVNet}&L &\scalebox{0.8}{[480,480,24]}&67.5&96.7&57.0&56.0&\textbf{72.6}&54.5&69.4&75.1&58.9&90.7&65.0&75.2&26.2&91.5&67.3&85.2&72.6&68.8&63.5&66.0&62\\

				2DPASS~\cite{LiDAR_2DPASS}&L &\scalebox{0.8}{[1000,1000,60]}&67.4&96.3&51.1&55.8&54.9&51.6&76.8&79.8&30.3&89.8&62.1&73.8&33.5&91.9&68.7&86.5&72.3&71.3&63.7&70.2&62/66*\\
				
				2DPASS~\cite{LiDAR_2DPASS}&L+C &\scalebox{0.8}{[1000,1000,60]}&72.9&97.0&\textbf{81.3}&63.4&61.1&61.5&77.9&81.3&\textbf{74.1}&89.7&67.4&74.7&40.0&\textbf{93.5}&\textbf{72.9}&86.2&73.9&71.0&65.0&70.4&62/66*\\

				GASN~\cite{LiDAR_GASN} &L&\scalebox{0.8}{[480,480,24]}&70.7&96.9&65.8&58.0&59.3&61.0&\textbf{80.4}&\textbf{82.7}&46.3&89.8&66.2&74.6&30.1&92.3&69.6&\textbf{87.3}&73.0&72.5&66.1&71.6&59\\

				RPVNet~\cite{LiDAR_RPVNet}&L &\scalebox{0.8}{[3200,3200,600]}&70.3&\textbf{97.6}&68.4&68.7&44.2&61.1&75.9&74.4&73.4&\textbf{93.4}&70.3&\textbf{80.7}&33.3&\textbf{93.5}&72.1&86.5&75.1&71.7&64.8&61.4&168/174*\\

				RangViT~\cite{LiDAR_RangeViT}&L&\scalebox{0.8}{[64,2048]}&64.0&95.4&55.8&43.5&29.8&42.1&63.9&58.2&38.1&93.1&70.2&80.0&32.5&92.0&69.0&85.3&70.6&71.2&60.8&64.7&-\\
				
				PVKD~\cite{LiDAR_PVKD}&L &\scalebox{0.8}{[480,360,32]}&71.2&97.0&67.9&69.3&53.5&60.2&75.1&73.5&50.5&91.8&70.9&77.5&41.0&92.4&69.4&86.5&73.8&71.9&64.9&65.8&76/90*\\

				WaffleIron~\cite{LiDAR_WafffeIron}&L&\scalebox{0.8}{[1000,1000,50]}&70.8&97.2&70.0&69.8&40.4&59.6&77.1&75.5&41.5&90.6&70.4&76.4&38.9&93.5&72.3&86.7&75.7&71.7&66.2&71.9&193\\
				
				LinK~\cite{LiDAR_LinK}&L&\scalebox{0.8}{[3200,3200,600]}&70.7&97.4&58.4&56.6&52.9&64.2&72.3&77.0&69.1&90.6&68.2&76.2&34.5&92.0&68.8&85.7&74.3&70.5&64.8&69.5&87/87*\\
				
				RangeFormer~\cite{LiDAR_RangeFormer}&L&\scalebox{0.8}{[64,2048]}&73.3&96.7&69.4&73.7&59.9&\textbf{66.2}&78.1&75.9&58.1&92.4&\textbf{73.0}&78.8&42.4&92.3&70.1&86.6&73.3&72.8&66.4&66.6&\textbf{37}\dag\\
				
				SFPNet~\cite{LiDAR_SFPNet}&L&\scalebox{0.8}{[1024,1024,180]}&70.3&97.2&64.9&63.8&44.8&54.7&70.4&74.6&52.9&91.9&70.6&78.0&39.7&93.3&71.5&85.4&73.7&70.1&66.1&\textbf{72.1}&-\\

				\hline

				Baseline&L&\scalebox{0.8}{[120,360,32]}&69.5&96.9&56.2&63.4&55.6&59.5&73.5&82.1&61.2&90.2&63.0&75.0&31.7&91.4&68.1&84.5&70.8&69.1&59.4&68.2&52\\

				Ours&L&\scalebox{0.8}{[120,360,32]}&\textbf{73.6}&97.4&70.9&\textbf{74.9}&60.2&63.6&79.2&77.6&54.6&91.6&71.5&77.7&\textbf{45.5}&92.8&70.9&86.9&\textbf{75.3}&\textbf{72.6}&\textbf{66.9}&68.9&56\\
				\hline

			\end{tabular}
		}
	\end{center}
	\caption{Results on SemanticKITTI test dataset. The \textbf{bold} numbers indicate the best results. `\textbf{L}' represents using LiDAR point cloud data as input. `\textbf{C}' represents using RGB data as input. `\textbf{*}' represents the inference speed which is measured by NVIDIA RTX 3090 for methods with open-source code and performance close to ours. \textbf{\dag} represents the inference speed is measured by NVIDIA A100. `Baseline' refers to the configuration without NUC and NUMA.}
	\label{tab:kitti}
\end{table*}
\textbf{SemanticKITTI} As shown in Table \ref{tab:kitti}, our method achieves state-of-the-art performance with significantly faster speed. The `baseline' in  Table~\ref{tab:kitti} refers to using an input resolution of $120\times360\times32$ without using the proposed non-uniform partition method. Compared with the uniform cylindrical partition counterpart Cylinder3D~\cite{LiDAR_Cylindrical}, our method achieves $5.8\%$ higher mIoU and $3 \times$ inference faster. This improvement stems from our approach, which not only reduces the input resolution to accelerate inference but also optimally adjusts the radial intervals to enhance segmentation performance. Our method achieves superior performance compared with the efficient LiDAR segmentation methods for both efficiency and performance, such as GASN~\cite{LiDAR_GASN}, DRINet~\cite{LiDAR_DRINet}, 2DPASS~\cite{LiDAR_2DPASS} and PVKD~\cite{LiDAR_PVKD}. Our method achieves comparable performance with SphereFormer~\cite{LiDAR_Spherical} and significantly faster by 2.19 times.

\begin{figure*}
	\begin{center}
		\centering
		
		\includegraphics[scale=0.26]{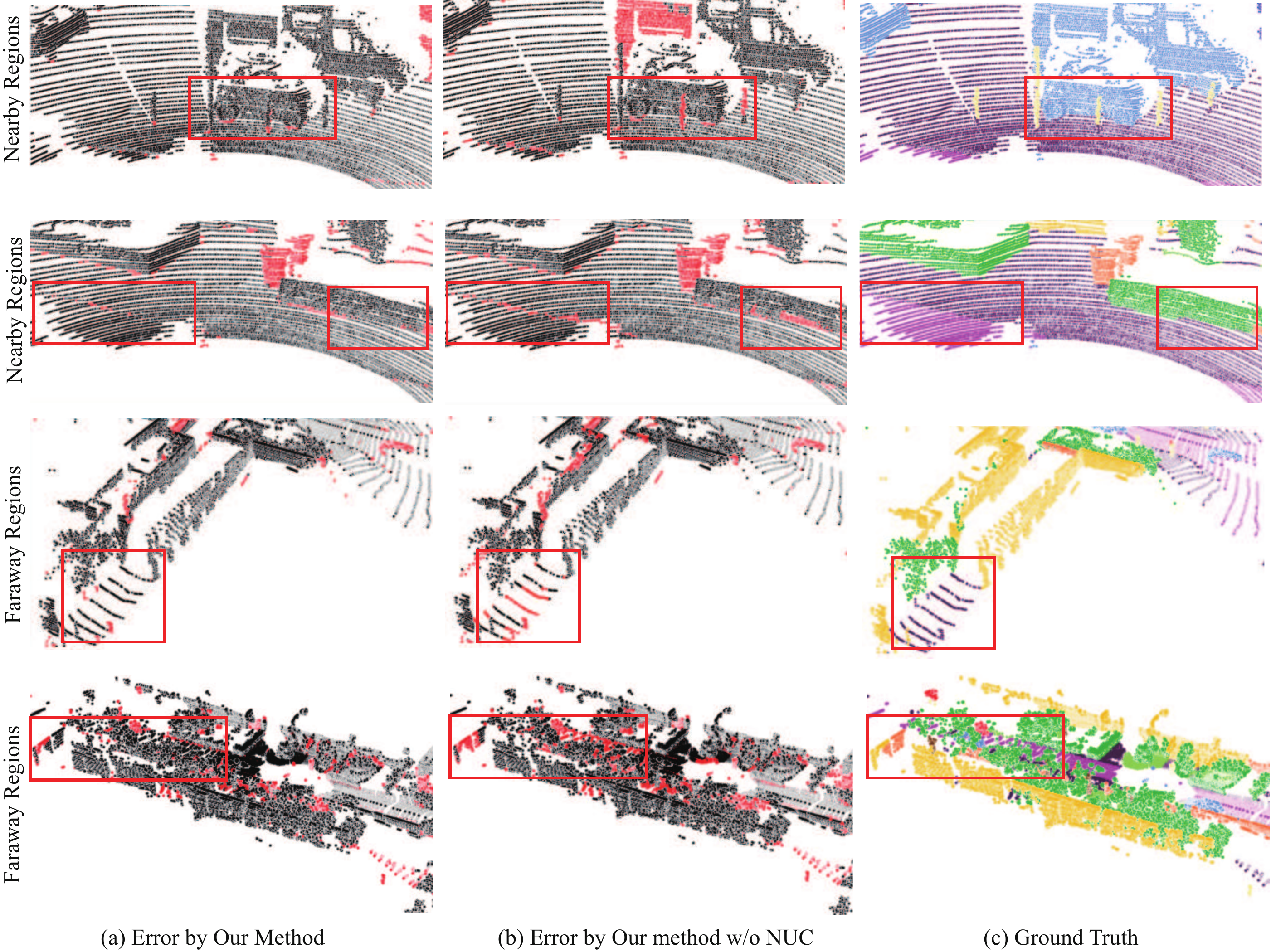}
	\end{center}
	\caption{Comparison between the results of our method with and w/o non-uniform cylindrical partition(NUC) on SemanticKITTI validation set.}

	\label{fig:vis}
	\vspace{-0.5cm}
	
\end{figure*}

The visualization results on SemanticKITTI are shown in Fig.~\ref{fig:vis}. We can observe that our method achieves superior performance in the nearby by reducing the encoding error and faraway regions by improving the receptive fields. As shown in the first and the second row, the errors in the boundary regions of the object decrease by reducing the encoding error. As shown in the third and the fourth row, the area with sparse point clouds in the faraway regions achieves better segmentation performance due to obtaining a larger receptive field.

The process of converting LiDAR points cloud to voxel representation using the maximum encoding strategy leads to encoding errors. A smaller radial interval in the nearby regions will decrease the encoding error.

\textbf{NuScenes} To verify the generalization ability of our method, we conduct experiments on the nuScenes dataset. Table \ref{tab:nuscenes_validation} shows the results on the nuScenes validation dataset following ~\cite{LiDAR_Cylindrical,LiDAR_PVKD,LiDAR_RPVNet}. Our method achieves state-of-the-art performance with faster speed and achieves the best performance for most of the categories.

\begin{table*}
	\setlength\tabcolsep{3pt}
	\tiny
	
	\begin{center}
		\resizebox{\linewidth}{!}{
			\begin{tabular}{c|c|cccccccccccccccc|c}
				\hline
				
				Methods&\rotatebox{90}{Mean IoU}&\rotatebox{90}{barrier}& \rotatebox{90}{bicycle}&\rotatebox{90}{bus}&\rotatebox{90}{car}&\rotatebox{90}{construction}&\rotatebox{90}{motorcycle}&\rotatebox{90}{pedestrian}&\rotatebox{90}{traffic-cone}&\rotatebox{90}{trailer}&\rotatebox{90}{truck}&\rotatebox{90}{driveable}&\rotatebox{90}{other}&\rotatebox{90}{sidewalk}&\rotatebox{90}{terrain}&\rotatebox{90}{manmade}&\rotatebox{90}{vegetation}&\rotatebox{90}{speed (ms)}\\
				\hline

				RangeNet++~\cite{LiDAR_Rangenet++}
				&65.5&66.0&21.3&77.2&80.9&30.2&66.8&69.6&52.1&54.2&72.3&94.1&66.6&63.5&70.1&83.1&79.8&-\\

				PolarNet
				~\cite{LiDAR_PolarNet} &71.0&74.7&28.2&85.3&90.9&35.1&77.5&71.3&58.8&57.4&76.1&96.5&71.1&74.7&74.0&87.3&85.7&-\\
				Salsanext
				~\cite{LiDAR_PolarNet} &72.2&74.8&34.1&85.9&88.4&42.2&72.4&72.2&63.1&61.3&76.5&96.0&70.8&71.2&71.5&86.7&84.4&-\\
				
				Cylinder3D
				~\cite{LiDAR_PolarNet} &76.1&76.4&40.3&91.2&93.8&51.3&78.0&78.9&64.9&62.1&84.4&96.8&71.6&76.4&75.4&90.5&87.4&98\\

				PVKD
				~\cite{LiDAR_PVKD} &76.0&76.2&40.0&90.2&\textbf{94.0}&50.9&77.4&78.8&64.7&62.0&84.1&96.6&71.4&76.4&76.3&90.3&86.9&54\\
				
				RPVNet
				~\cite{LiDAR_RPVNet} &77.6&\textbf{78.2}&43.4&92.7&93.2&49.0&85.7&80.5&66.0&66.9&84.0&96.9&73.5&75.9&76.0&90.6&88.9&-\\
				
				RangeViT
				~\cite{LiDAR_RangeViT} &75.2&75.5&40.7&88.3&90.1&49.3&79.3&77.2&66.3&65.2&80.0&96.4&71.4&73.8&73.8&89.9&87.2&-\\
				
				RangeFormer
				~\cite{LiDAR_RangeFormer} &78.1&78.0&45.2&94.0&92.9&58.7&83.9&77.9&69.1&63.7&85.6&96.7&74.5&75.1&75.3&89.1&87.5&-\\
				
				WaffleIron
				~\cite{LiDAR_WafffeIron} &79.1&79.8&\textbf{53.8}&94.3&87.6&49.6&\textbf{89.1}&\textbf{83.8}&\textbf{70.6}&72.7&84.9&\textbf{97.1}&75.8&76.5&75.9&87.8&86.3&92\\
				
					SFPNet
				~\cite{LiDAR_SFPNet} &\textbf{80.1}&78.8&49.7&\textbf{95.3}&93.5&\textbf{63.1}&86.4&82.9&68.6&\textbf{72.8}&\textbf{86.7}&97.0&74.7&76.0&75.3&\textbf{91.2}&\textbf{89.5}&-\\
				
				\hline

				Baseline&74.7&73.4&43.0&89.3&85.1&44.7&83.1&77.5&65.9&64.9&80.5&94.7&75.8&73.9&73.7&86.3&85.4&\textbf{33}\\

				Ours&78.9&78.1&48.6&92.9&92.8&48.5&87.2&81.4&\textbf{70.6}&70.9&83.9&\textbf{97.1}&\textbf{78.7}&\textbf{77.6}&\textbf{76.4}&88.9&88.6&35\\
				\hline

			\end{tabular}
		}
	\end{center}
	\caption{Quantitative results of the proposed method and state-of-the-art LiDAR segmentation methods on nuScenes validation following existing works~\protect\cite{LiDAR_Cylindrical,LiDAR_PVKD,LiDAR_RPVNet,LiDAR_RangeViT}.}
	\label{tab:nuscenes_validation}
\end{table*}

\begin{table*}
	\setlength\tabcolsep{3pt}
	\tiny
	
	\begin{center}
		\resizebox{\linewidth}{!}{
			\begin{tabular}{c|c|c|cccccccccccccccc|c}
				\hline
				
				Methods&Input&\rotatebox{90}{Mean IoU}&\rotatebox{90}{barrier}& \rotatebox{90}{bicycle}&\rotatebox{90}{bus}&\rotatebox{90}{car}&\rotatebox{90}{construction}&\rotatebox{90}{motorcycle}&\rotatebox{90}{pedestrian}&\rotatebox{90}{traffic-cone}&\rotatebox{90}{trailer}&\rotatebox{90}{truck}&\rotatebox{90}{driveable}&\rotatebox{90}{other}&\rotatebox{90}{sidewalk}&\rotatebox{90}{terrain}&\rotatebox{90}{manmade}&\rotatebox{90}{vegetation}&\rotatebox{90}{speed(ms)}\\
				\hline
				
				PolarNet~\cite{LiDAR_PolarNet} &L&69.4&72.2&16.8&77.0&86.5&51.1&69.7&64.8&54.1&69.7&63.5&96.6&67.1&77.7&72.1&87.1&84.5&-\\
				
				JS3C-Net~\cite{LiDAR_JS3C-Net} &L&73.6&80.1&26.2&87.8&84.5&55.2&72.6&71.3&66.3&76.8&71.2&96.8&64.5&76.9&74.1&87.5&86.1&-\\

				Cylinder3D~\cite{LiDAR_Cylindrical} &L&77.2&82.8&29.8&84.3&89.4&63.0&79.3&77.2&73.4&84.6&69.1&\textbf{97.7}&70.2&\textbf{80.3}&75.5&90.4&87.6&106\\

				AMVNet~\cite{LiDAR_AMVNet} &L&77.3&80.6&32.0&81.7&88.9&67.1&84.3&76.1&73.5&84.9&67.3&97.5&67.4&79.4&75.5&91.5&88.7&85\\
				SPVCNN~\cite{LiDAR_NAS} &L&77.4&80.0&30.0&91.9&90.8&64.7&79.0&75.6&70.9&81.0&74.6&97.4&69.2&80.0&76.1&89.3&87.1&110\\

				AF2S3Net~\cite{LiDAR_AF-S3Net} &L&78.3&78.9&52.2&89.9&84.2&\textbf{77.4}&74.3&77.3&72.0&83.9&73.8&97.1&66.5&77.5&74.0&87.7&86.8&270\\

				2DPASS~\cite{LiDAR_2DPASS} &L&77.6&80.8&37.9&92.7&90.5&65.4&77.6&71.5&70.9&83.1&75.3&97.0&69.3&78.1&75.6&89.1&86.8&44\\
				PMF~\cite{LiDAR_PMF} &L+C&77.0&82.0&40.0&81.0&88.0&64.0&79.0&80.0&\textbf{76.0}&81.0&67.0&97.0&68.0&78.0&74.0&90.0&88.0&125\\
				
				2D3DNet~\cite{LiDAR_2D3DNet} &L+C&80.0&83.0&\textbf{59.4}&88.0&85.1&63.7&84.4&\textbf{82.0}&\textbf{76.0}&84.8&71.9&96.9&67.4&79.8&76.0&92.1&89.2&-\\
				
				RangeFormer~\cite{LiDAR_RangeFormer} &L&80.1&\textbf{85.6}&47.4&91.2&90.9&70.7&84.7&77.1&74.1&83.2&72.6&97.5&70.7&79.2&75.4&91.3&88.9&-\\

				SFPNet~\cite{LiDAR_SFPNet} &L&\textbf{80.2}&83.7&42.5&89.1&91.5&74.1&83.5&79.1&74.7&\textbf{87.3}&73.3&\textbf{97.7}&\textbf{78.1}&\textbf{80.3}&\textbf{76.2}&\textbf{92.3}&\textbf{89.3}&-\\
				
				\hline

				Baseline&L&76.1&78.3&31.1&89.1&86.8&70.2&76.7&74.9&67.8&83.6&75.7&97.0&65.7&76.0&73.9&87.1&84.2&\textbf{33}\\
				
				Ours&L&\textbf{80.2}&82.9&39.2&\textbf{94.6}&\textbf{91.6}&75.6&\textbf{85.6}&77.3&73.3&85.1&\textbf{76.8}&97.6&70.5&80.1&75.5&89.5&86.9&35\\
				\hline

			\end{tabular}
		}
	\end{center}
	\caption{Quantitative results of the proposed method and state-of-the-art LiDAR segmentation methods on nuScenes test set.}
	\label{tab:nuscenes_test}
\end{table*}

As shown in  Tab.~\ref{tab:nuscenes_test}, our method achieves state-of-the-art performance with much faster speed on nuScenes test set. `Baseline' represents using the uniform partition method, while `Ours' represents using the proposed non-uniform partition method. `L' represents using LiDAR point cloud as input and `C' represents using camera image data as input. PMF~\cite{LiDAR_PMF} and 2D3DNet~\cite{LiDAR_2D3DNet} use the additional RGB data to boost the performance, while our method only uses the LiDAR point cloud data.

Note that in Table~\ref{tab:nuscenes_validation} and Table~\ref{tab:nuscenes_test} we only include the published works and the results are from their corresponding paper. We did not compare our method with SPVCNN++ as the details of the method are unavailable.  And we did not compare with LidarMultiNet~\cite{LiDAR_LidarMultiNet} as it uses the past 9 point clouds shown in their paper, while our method only uses the current point cloud.

\subsection{Ablation Study}

\begin{table*}
	\small
	\begin{center}	
		\begin{tabular}{c c c  c|c c c c}
			\hline
			Baseline  &   NUC  & NUMA & IA &mIoU(\%)& Param(M)&MACs(G)&Speed(ms)  \\
			\hline
			
			\checkmark	&              &   &     &63.2 &\textbf{49.7}&\textbf{19.1}&\textbf{52}  \\
			\checkmark	&       \checkmark    &      &   & 67.6 &\textbf{49.7}&19.9&55 \\
			\checkmark	&         \checkmark    &       \checkmark   &    &  68.3 &49.9&19.9&56  \\
			\checkmark	&        \checkmark    &       \checkmark   &   \checkmark &  \textbf{70.3}&49.9&19.9&56   \\

			\hline
		\end{tabular}
		
	\end{center}
	
	\caption{Ablation study for the core components of NUC-Net on SemanticKITTI validation set.}
	\label{tab:core_component}
\end{table*}

\textbf{The Effectiveness of the Proposed Components}
In this section, we conduct ablation studies on the key components of NUC-Net. The experimental results are shown in Table \ref{tab:core_component}. `Baseline' means that uses the 3D cylindrical coordinates and the 3D convolution network. In other words, the baseline method is derived by removing the NUMA and IA modules from the NUC-Net and replacing the non-uniform cylindrical partition with a uniform cylindrical partition.  `NUC' represents the proposed non-uniform cylindrical partition method. `NUMA' represents the proposed non-uniform multi-scale aggregation method. `IA' represents instance augmentation. The proposed non-uniform partition can obtain an improvement by $4.4\%$ mIoU.  The multi-scale aggregation method obtains the performance gain by $0.7\%$ mIoU. Instance augmentation boosts the performance by $2.0\%$ mIoU.

\begin{table}
	\setlength\tabcolsep{3pt}
	\footnotesize
	
\begin{center}	
	\begin{tabular}{c|c|c|c}
		\hline
		
		Methods&Partition Mechanism&Voxel Resolution&{mIoU}\\
		\hline																				
		\multirow{6}*{PolarNet
			~\cite{LiDAR_PolarNet}}&\multirow{3}*{Uniform}&[120,180,32] &52.1\\
		&&[120,360,32] &52.6\\
		
		&&[480,360,32]&54.9 \\
		
		&\multirow{3}*{Non-uniform (API)}&[120,180,32]&57.6 (\textbf{+5.5}) \\
		
		&&[120,360,32]&58.7 (\textbf{+6.1}) \\
		&&[480,360,32]&59.6 (\textbf{+4.7}) \\ 
		\hline				
		\multirow{6}*{Cylinder3D
			~\cite{LiDAR_Cylindrical}}&\multirow{3}*{Uniform}&[120,180,32] &59.8\\
		&&[120,360,32] &61.8\\
		
		&&[480,360,32]& 64.5\\
		
		&\multirow{3}*{Non-uniform (API)}&[120,180,32]&65.6 (\textbf{+5.8}) \\
		
		&&[120,360,32]& 66.5 (\textbf{+4.7})\\
		&&[480,360,32]&66.8 (\textbf{+2.3}) \\
		
		\hline
	\end{tabular}
	\end{center}
	\caption{Results of before and after exploiting the non-uniform partition method on PolarNet and Cylinder3D. `Uniform' represents the uniform partition methods proposed by PolarNet~\protect\cite{LiDAR_PolarNet} or Cylinder3D~\protect\cite{LiDAR_Cylindrical}. `Non-uniform' represents the proposed non-uniform partition(Arithmetic Progression of Interval) method. }
	\label{tab:generality}
	\vspace{-0.5cm}
\end{table}

\textbf{Generality of Non-Uniform Partition}
To evaluate the generality of the non-uniform partition mechanism, we do ablation study by adding the proposed non-uniform partition mechanism to PolarNet~\cite{LiDAR_PolarNet} and Cylinder3D~\cite{LiDAR_Cylindrical}. We replace the partition methods of PolarNet and Cylinder3D with the proposed non-uniform partition mechanism while keeping other setting unchanged as their open-source codes. The proposed non-uniform partition mechanism boosts the performance of Cylinder3D by $5.8\%$ and the performance of PolarNet by $6.1\%$ at most. Demonstrating its capability as a `model-independent' and `resolution-independent' mechanism, our non-uniform partition mechanism proves to be a general component for LiDAR semantic segmentation.

\begin{figure}
	
	\centering
	\includegraphics[width=7cm]{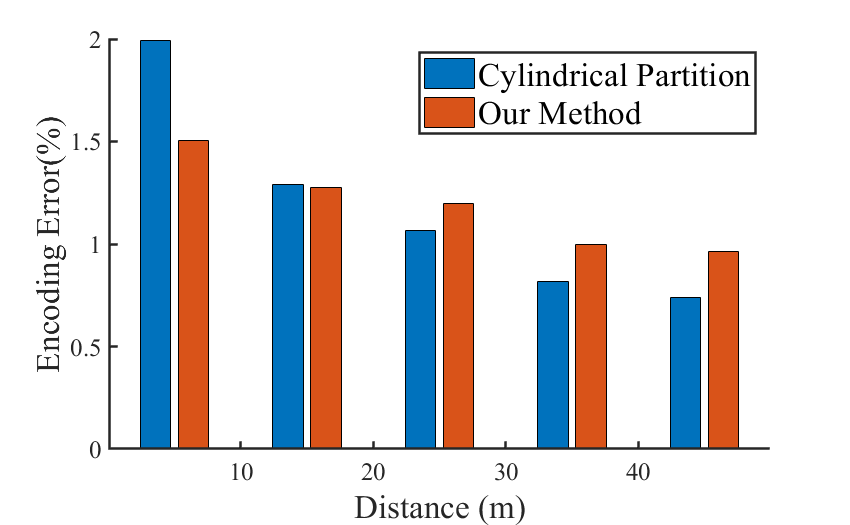} 
	\caption{The encoding errors with and w/o non-uniform partition on SemanticKITTI training set with different distance-range.}
	\label{fig:encoding_error_distance}
\end{figure}

\begin{figure}
	\centering
	\includegraphics[width=7cm]{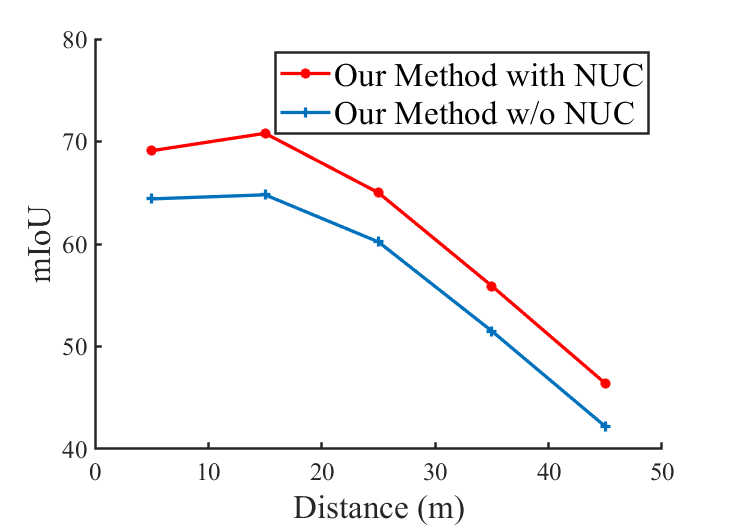} 
	\caption{The results on SemanticKITTI validation set with different distance-range.}
	\label{fig:miou_distance}
	\vspace{-0.5cm}
\end{figure}
\textbf{Design Choices of Non-Uniform Partition}
Table \ref{tab:design_choice} shows different design choices of non-uniform partition mechanisms. `Baseline' represents that uses the uniform cylindrical partition method. `API' represents that uses the Arithmetic Progression of Interval to partition the point cloud. `GPI' represents that uses the Geometric Progression of Interval to partition the point cloud. For the Geometric Progression of Interval, each interval after the first is found by multiplying the previous interval by the common ratio, where $a_0$ denotes the first term and $r$ denotes the common ratio. `Piecewise' means that we use different fixed intervals in different radial ranges. The piecewise partition strategy divides the scene based on radial distance into three parts: near distance region (0-15m), middle distance region (15-30m), and far distance region (30-50m). Different numbers of partitions are then applied to radial distance.  We divide the radial direction into 120 partitions and apply different numbers of divisions to the near, middle, and far regions. The ratios of the three regions are shown in parentheses in the table. For example, (7,3,2) represents dividing the near distance region into 70 partitions, the middle distance region into 30 partitions, and the far distance region into 20 partitions. In the `increasing d' method, d gradually grows, and d' represents the amount by which d increases as the radial index rises. We have tried a series of reasonable parameters for each comparison partition mechanism and use the best one in Table \ref{tab:design_choice}.

\begin{table}
	 
	\footnotesize
	\begin{center}
		\footnotesize
		\begin{tabular}{c c c}
			\hline
			Method&Parameter Condidates &mIoU(\%)  \\
			\hline
			Baseline     & - &      65.5 \\	
			\hline
			\multirow{4}*{Piecewise
			}          &   (7,3,2)   &  67.2  \\
			& (8,3,1)    &   67.5  \\
			&  (7,4,1)   &   66.5  \\
			&  (5,4,3)   &   65.7  \\
			\hline
			\multirow{3}*{Increasing `d'}
			& $a^0=0.05, d=0.0050,d'=0.000030$     & 69.1 \\
			& $a^0=0.05, d=0.0052,d'=0.000025$       &69.2  \\
			& $a^0=0.05, d=0.0054,d'=0.000020$     & 68.4  \\
			
			\hline
			\multirow{4}*{GPI}
			& $a^0=0.05, r=1.0541$       &66.9  \\
			& $a^0=0.1, r=1.0465$      & 68.6  \\
			&  $a^0=0.2, r=1.0391$      &67.8  \\
			& $a^0=0.3, r=1.0345$       &67.7  \\
			
			\hline
			\multirow{3}*{API (Ours)}          
			& $a^0=0.04, d=0.0064$       &68.1  \\
			& $a^0=0.05, d=0.0062$      & \textbf{70.3}  \\
			&  $a^0=0.06, d=0.0060$      &68.8  \\
			
			\hline
			
		\end{tabular}
	\end{center}
	\caption{Different non-uniform partition mechanism on SemanticKITTI validation set.}
	\label{tab:design_choice}
\end{table}

As shown in  Table \ref{tab:design_choice}, our baseline method only achieves a lower result of 65.5 mIoU. All of the non-uniform partition mechanisms can boost the performance of LiDAR segmentation. `API' outperforms baseline method by 4.8 mIoU. `API' achieves the best result as `API' is more consistent with the point density variation of the point cloud.    


\textbf{Different Settings for API} In this section, we analyze different design choices for the Arthmetric Progression of Interval on SemanticKITTI and nuScenes. We do ablation study using different first terms and tolerances. The experiment is conducted by using an input voxel resolution of $120 \times 360 \times 32$. As shown in Table \ref{tab:API}, we obtain the best performance by using first term$=0.05$ and tollerance=$0.0062$. And all the parameter settings achieve consistent improvement compared with the uniform-partition method on SemanticKITTI~\cite{dataset_semanticKITTI} and nuScenes dataset~\cite{dataset_nuScenes}. 

Since the majority of point clouds are concentrated in the nearby region (0-15m), parameters that produce smaller radial intervals in the nearby region are preferred, resulting in a lower point cloud density. Take SemanticKITTI as an example, we first analyze the point cloud density at different distances in the SemanticKITTI dataset and observe that the highest point cloud density occurs around the 7m region, gradually decreasing on both sides of 7m. We then calculate the radial interval of voxels at 7m under different values of $a_0$ and $d$. Within the parameter range of $0.04 \leq a_0 \leq 0.16$, the voxel radial intervals in the 7m region are relatively smaller for various parameter choices, all being less than 0.3m, and the intervals are quite close to each other. Then, considering that larger radial intervals in the faraway region could boost LiDAR semantic segmentation, we prefer smaller values of $a_0$ within the parameter range of $0.04 \leq a_0 \leq 0.16$. After that, we select several parameter settings around $a_0 = 0.05$ and train them individually on the training set. We then choose the parameter that achieves the best LiDAR point cloud segmentation performance on the validation set as the final parameter.

\begin{table}

	\begin{center}
	\small
	\begin{tabular}{c|ccc}
		\hline
		
		Dataset&Parameter Condidates&{mIoU}\\
		\hline

		\multirow{3}*{SemanticKITTI
		}
		& $a^0$=0.04, d=0.0064&68.1 \\
		&$a^0$=0.05, d=0.0062&\textbf{70.3} \\
		
		&$a^0$=0.06, d=0.0060&68.8\\
		
		\hline
		\multirow{4}*{nuScenes
		}
		& $a^0$=0.04, d=0.0064&77.8 \\
		&$a^0$=0.05, d=0.0062&\textbf{78.9} \\
		
		&$a^0$=0.06, d=0.0060&78.7\\
		&$a^0$=0.07, d=0.0058&77.6\\

		\hline
		
	\end{tabular}
	
	\end{center}
	\caption{Results of different parameters for Arithmetic Progression of Interval Method(API) on SemanticKITTI and nuScenes validation set. $a^0$ represents the first term and $d$ represents the tolerance. }
	\label{tab:API}
\end{table}
\begin{table*}

	\begin{center}
		\small
		\begin{tabular}{c|ccccc}
			\hline
			
			Method&Voxel Resolution&{mIoU}&Param(M)&MACs(G)&Speed(ms)\\
			\hline

			\multirow{3}*{NUC-Net
			}&[120,180,32]&69.4&49.9 &\textbf{12.7}&
			\textbf{49}\\
			&[120,360,32]&70.3&49.9 &19.9&56\\
			
			&[480,360,32]&\textbf{70.4}&49.9&59.8&135\\

			\hline
			
		\end{tabular}
		
	\end{center}
	\caption{Results of model efficiency and accuracy of different input voxel resolutions on SemanticKITTI validation set.}
	\label{tab:resolution}
\end{table*}

	\begin{table*}
	\setlength\tabcolsep{3pt}
	\tiny
	
	\begin{center}
		\resizebox{\linewidth}{!}{
			\begin{tabular}{c|c|ccccccccccccccccccccccccc}
				\hline
				
				Methods&\rotatebox{90}{Mean IoU}&\rotatebox{90}{Car}& \rotatebox{90}{Bicycle}&\rotatebox{90}{Motorcycle}&\rotatebox{90}{Truck}&\rotatebox{90}{Other-vehicle}&\rotatebox{90}{Person}&\rotatebox{90}{Bicyclist}&\rotatebox{90}{Motorcyclist}&\rotatebox{90}{Road}&\rotatebox{90}{Parking}&\rotatebox{90}{Sidewalk}&\rotatebox{90}{Other-ground}&\rotatebox{90}{Building}&\rotatebox{90}{Fence}&\rotatebox{90}{Vegetation}&\rotatebox{90}{Trunk}&\rotatebox{90}{Terrain}&\rotatebox{90}{Pole}&\rotatebox{90}{traffic sign}&\rotatebox{90}{mov. car}&\rotatebox{90}{mov. bicylist}&\rotatebox{90}{mov. person}&\rotatebox{90}{mov. motorcyc.}&\rotatebox{90}{mov. truck}&\rotatebox{90}{mov. other veh.}\\
				\hline

				TangentConv~\cite{LiDAR_TangentConv} &34.1&84.9&2.0&18.2&21.1&18.5&1.6&0.0&0.0&83.9&38.3&64.0&15.3&85.8&49.1&79.5&43.2&56.7&36.4&31.2&40.3&1.1&6.4&1.9&20.1&42.2\\

				DarkNet53~\cite{dataset_semanticKITTI}&41.6&84.9&2.0&18.2&21.1&18.5&1.6&0.0&0.0&83.9&38.3&64.0&15.3&85.8&49.1&78.4&50.7&64.8&38.1&53.3&61.5&14.1&15.2&0.2&28.9&37.8\\

				SpSeqnet~\cite{SpSequenceNet}&43.1&88.5&24.0&26.2&29.2&22.7&6.3&0.0&0.0&90.1&57.6&73.9&27.1&91.2&\textbf{66.8}&84.0&66.0&65.7&50.8&48.7&53.2&41.2&26.2&36.2&2.3&0.1\\
				
				KPConv~\cite{LiDAR_KPConv}&51.2&93.7&44.9&47.2&42.5&38.6&21.6&0.0&0.0&86.5&58.4&70.5&26.7&90.8&64.5&84.6&70.3&66.0&57.0&53.9&69.4&0.5&0.5&\textbf{67.5}&\textbf{67.4}&\textbf{47.2}\\

				Cylinder3D~\cite{LiDAR_Cylindrical}&51.5&93.8&\textbf{67.6}&\textbf{63.3}&41.2&37.6&12.9&\textbf{0.1}&\textbf{0.1}&\textbf{90.4}&\textbf{66.3}&\textbf{74.9}&\textbf{32.1}&\textbf{92.4}&65.8&\textbf{85.4}&\textbf{72.8}&68.1&\textbf{62.6}&61.3&68.1&0.0&0.1&63.1&60.0&0.4\\
				\hline
				
				Ours&\textbf{53.8}&\textbf{95.1}&53.9&55.6&\textbf{42.6}&\textbf{43.6}&\textbf{22.5}&0.0&0.0&90.0&61.9&74.0&26.0&91.7&66.5&85.2&71.0&\textbf{69.5}&60.3&\textbf{69.2}&\textbf{78.6}&\textbf{66.6}&\textbf{69.1}&35.5&1.8&15.9\\
				\hline

			\end{tabular}
		}
	\end{center}
	\caption{The experimental results on SemantiKITTI multi-scan benchmark. The \textbf{bold} numbers indicate the best results.}
	\label{tab:kitti-multi-scan}
\end{table*}
\begin{table*}
	\setlength\tabcolsep{3pt}
	\footnotesize
	
	\begin{center}
			\begin{tabular}{c|c|ccc|ccc|cccc}
				\hline
				
				Methods&PQ&PQ$^{\dag}$&RQ&SQ&PQ$^{\rm Th}$&RQ$^{\rm Th}$&SQ$^{\rm Th}$&PQ$^{\rm St}$&RQ$^{\rm St}$&SQ$^{\rm St}$&mIoU\\
				\hline

				KPConv~\cite{LiDAR_KPConv}+PV-RCNN~\cite{dection_PV-RCNN}
				&51.7&57.4&63.1&\textbf{78.9}&46.8&56.8&81.5&\textbf{55.2}&\textbf{67.8}&\textbf{77.1}&63.1\\

				PointGroup++
				~\cite{Panoptic_PointGroup} &46.1&54.0&56.6&74.6&47.7&55.9&73.8&45.0&57.1&75.1&55.7\\

				LPASD
				~\cite{Panoptic_LPASD}  &36.5&46.1&-&-&-&28.2&-&-&-&-&50.7\\

				Cylinder3D
				~\cite{LiDAR_Cylindrical} &56.4&62&67.1&76.5&58.8&66.8&75.8&54.8&67.4&\textbf{77.1}&63.5\\
				\hline
				Baseline &54.7&59.4&65.3&76.5&58.9&67.9&85.2&51.6&63.4&70.1&59.9\\
				
				Ours &\textbf{57.9}&\textbf{63.5}&\textbf{68.7}&77.9&\textbf{63.2}&\textbf{72.0}&\textbf{87.3}&54.1&66.3&70.9&\textbf{63.9}\\
				\hline

			\end{tabular}
	\end{center}
	\caption{LiDAR panoptic segmentation results on the validation set of SemanticKITTI. The \textbf{bold} numbers indicate the best results. }
	\label{tab:panoptic}
	\vspace{-0.5cm}
\end{table*}

\begin{table*}
	\setlength\tabcolsep{3pt}
	\tiny
	
	\begin{center}
		\resizebox{\linewidth}{!}{
			\begin{tabular}{c|c|ccccccccccccccccccc}
				\hline
				
				Methods&\rotatebox{90}{Mean IoU}&\rotatebox{90}{Car}& \rotatebox{90}{Bicycle}&\rotatebox{90}{Motorcycle}&\rotatebox{90}{Truck}&\rotatebox{90}{Other-vehicle}&\rotatebox{90}{Person}&\rotatebox{90}{Bicyclist}&\rotatebox{90}{Motorcyclist}&\rotatebox{90}{Road}&\rotatebox{90}{Parking}&\rotatebox{90}{Sidewalk}&\rotatebox{90}{Other-ground}&\rotatebox{90}{Building}&\rotatebox{90}{Fence}&\rotatebox{90}{Vegetation}&\rotatebox{90}{Trunk}&\rotatebox{90}{Terrain}&\rotatebox{90}{Pole}&\rotatebox{90}{traffic sign}\\
				\hline

				Cylinder3D~\cite{LiDAR_Cylindrical}+scribble~\cite{LiDAR_Scribble} &61.3&91.0&41.1&58.1&\textbf{85.5}&57.1&\textbf{71.7}&80.9&\textbf{0.0}&87.2&35.1&\textbf{74.6}&\textbf{3.3}&88.8&51.5&86.3&68.0&70.7&63.4&49.5\\
				
				MinkowskiNet~\cite{LiDAR_MinkowskiNet}+scribble~\cite{LiDAR_Scribble} &58.5&91.1&23.8&\textbf{59.0}&66.3&58.6&65.2&75.2&\textbf{0.0}&83.8&36.1&72.4&0.7&90.2&51.8&86.7&68.5&\textbf{72.5}&62.5&46.6\\

				SPVCNN~\cite{LiDAR_PVCNN}+scribble~\cite{LiDAR_Scribble} &60.8&91.1&35.3&57.2&71.1&\textbf{63.8}&70.0&81.3&\textbf{0.0}&84.6&37.9&72.9&0.0&90.0&\textbf{54.0}&87.4&71.1&73.0&\textbf{64.0}&50.5\\
				\hline
				
				Ours+scribble~\cite{LiDAR_Scribble} &\textbf{62.3}&\textbf{96.7}&\textbf{42.6}&58.9&83.6&56.1&71.0&\textbf{82.4}&\textbf{0.0}&\textbf{88.1}&\textbf{38.9}&72.9&2.1&\textbf{90.7}&53.1&\textbf{87.9}&\textbf{72.3}&71.5&63.6&\textbf{50.8}\\

				\hline

			\end{tabular}
		}
	\end{center}
	\caption{The performance of  LiDAR semantic segmentation on SemanticKITTI validation set using scribble annotations. The \textbf{bold} numbers indicate the best results. }
	\label{tab:scribble-kitti}
	\vspace{-0.5cm}
\end{table*}

\begin{table*}
	\small
	\begin{center}	
		\begin{tabular}{c| c| c  c c c c c c c }
			\hline
			Method  &   mCE$\downarrow$  & Fog & Wet &Snow&Move&Beam&Cross&Echo&Sensor  \\
			\hline
			
			MinkU$_{18}$~\cite{LiDAR_MinkowskiNet}&100&100&100&100&100&100&100&100&100   \\
			\hline
			SqSeg~\cite{LiDAR_SqSeg}&164.9&183.9&158.0&165.5&122.4&171.7&188.1&158.7&170.8   \\
			
			SqSegV2~\cite{LiDAR_SqueezeSegV2}&152.5&168.5&141.2&154.6&115.2&155.2&176.0&145.3&163.5   \\
			
			RGNet$_21$~\cite{LiDAR_Rangenet++}&136.3&156.3&128.5&133.9&102.6&141.6&148.9&128.3&150.6   \\
			
			RGNet$_54$~\cite{LiDAR_Rangenet++}&130.7&144.3&123.7&128.4&104.2&135.5&129.4&125.8&153.9   \\
			
			SalsaNext~\cite{LiDAR_Salsanext}&116.1&147.5&112.1&116.6&77.6&115.3&143.5&114.0&102.5   \\
			
			FIDNet~\cite{LiDAR_FIDNet}&113.8&127.7&105.1&107.7&88.9&116.0&121.3&113.7&130.0   \\
			
			CENet~\cite{LiDAR_CENet}&103.4&129.8&92.7&99.2&70.5&101.2&131.1&102.3&100.4   \\
			\hline
			PolarNet~\cite{LiDAR_PolarNet}&118.6&138.8&107.1&108.3&86.8&105.1&178.1&112.0&112.3   \\
			\hline
			KPConv~\cite{LiDAR_KPConv}&99.5&103.2&91.9&98.1&110.7&97.6&111.9&97.3&85.4   \\
			PIDS$_{1.2 \times}$~\cite{LiDAR_PIDS}&104.1&118.1&98.9&109.5&114.8&103.2&103.9&97.0&87.6   \\
			PIDS$_{2.0 \times}$~\cite{LiDAR_PIDS}&101.2&110.6&95.7&104.6&115.6&98.6&102.2&97.5&84.8   \\
			Waffle~\cite{LiDAR_WafffeIron}&109.5&123.5&90.1&108.5&99.9&93.2&186.1&\textbf{91.0}&84.1   \\
			\hline
			MinkU$_{34}$~\cite{LiDAR_MinkowskiNet}&100.6&105.3&99.4&106.7&98.7&97.6&99.9&99.0&98.3   \\
			Cylinder3D$_{SPC}$~\cite{LiDAR_Cylindrical}&103.3&142.5&92.5&113.6&70.9&97.0&105.7&104.2&99.7   \\
			Cylinder3D$_{RSC}$~\cite{LiDAR_Cylindrical}&103.1&142.5&101.3&116.9&61.7&98.9&111.4&99.0&93.4   \\
			\hline
			SPV$_{18}$~\cite{LiDAR_PVCNN}&100.3&101.3&100.0&104.0&97.6&99.2&100.6&99.6&100.2   \\
			SPV$_{34}$~\cite{LiDAR_PVCNN}&\textbf{99.2}&\textbf{98.5}&100.7&102.0&97.8&99.0&98.4&98.8&98.1   \\
			RPVNet~\cite{LiDAR_RPVNet}&111.7&118.7&101.0&104.6&78.6&106.4&185.7&99.2&99.8   \\
			CPGNet~\cite{LiDAR_CPGNet}&107.3&141.0&92.6&104.3&\textbf{61.1}&\textbf{90.9}&195.6&95.0&\textbf{78.2}\\
			
			2DPASS~\cite{LiDAR_2DPASS}&106.1&134.9&85.5&110.2&62.9&94.4&171.7&96.9&92.7\\
			
			GFNet~\cite{LiDAR_GFNet}&108.7&131.3&94.4&\textbf{92.7}&61.7&98.6&198.9&98.2&93.6\\
			\hline
			
			Cylinder3D$_{[120,360,32]}$~\cite{LiDAR_Cylindrical}&115.8&143.5&105.7&114.4&74.9&108.0&128.7&112.0&139.3\\
			Ours$_{[120,360,32]}$&99.9&119.0&\textbf{78.4}&104.0&73.7&94.7&\textbf{96.9}&105.1&127.4\\

			\hline
		\end{tabular}
		
	\end{center}
	
	\caption{Results of Corruption Error (CE) on SemanticKITTI-C. The \textbf{bold} numbers indicate the best results.}
	\label{tab:robo3D}
	\vspace{-8pt} 
\end{table*}

\textbf{Different Voxel Resolutions for API} Uniform partition methods acquire high-resolution voxel to preserve the fine details of the 3D scene. These methods usually suffer from low resolution. We do ablation studies for different voxel resolutions for the non-uniform partition method. As shown in Table \ref{tab:resolution}, with $4 \times$ resolution reduction for the radial axis, the performance of $120 \times 360 \times 32$ version of our method is on par with $480 \times 360 \times 32$ version and achieves $3 \times$ MACs reduction. And the performances of NUC's different resolution versions are significantly better than the uniform counterparts. Different from existing efficient LiDAR semantic segmentation methods using range/BEV maps~\cite{LiDAR_Rangenet++,LiDAR_PolarNet,LiDAR_RPVNet}, knowledge distillation~\cite{LiDAR_PVKD} and lightweight architecture~\cite{LiDAR_DRINet,LiDAR_RandLA-Net}, the proposed NUC-Net speeds up by the more efficient and representative NUC representation, which shows superior performance and complementary to existing efficient methods.
\begin{table*}
	\small
	\begin{center}
		\begin{tabular}{c c c c c c|c}
			\hline
			&  0-10m   &10-20m  &20-30m & 30-40m &40-50m&overall  \\
			\hline
			Ours ($120 \times 360 \times 32$)& 9516.8   &6662.4        &    \textbf{2719.8}   &   \textbf{1370.2}          &   \textbf{745.9}    & 21015.1  \\
			Uniform ($120 \times 360 \times 32$)	& \textbf{7525.2}      &  \textbf{6642.1}          &2999.3   &1574.1  & 873.1  &\textbf{19613.8}   \\
			Uniform ($480 \times 360 \times 32$)	&   15914.6     &    11714.4   &  4927.3    &2366.7 &1250.0  & 36173.1  \\

			\hline
		\end{tabular}
	\end{center}
	\caption{The number of non-empty voxels with different distance-range. }
	
	\label{tab:non_empty}
	\vspace{-8pt} 
\end{table*}

\begin{table*}
	\footnotesize
	\begin{center}
		\begin{tabular}{c c c c c c c c}
			\hline
			Method  &  Param(M) &MACs(G) & Speed(ms)  & Training Time(GPU-days) &GPU Memory(G)&mIoU(\%)  \\
			\hline

			Cylinder3D~\cite{LiDAR_Cylindrical}	&   53.1   & 63.8  &   171   &  10 & 6.7 / 3.0  &    67.8  \\

			RandLA-Net~\cite{LiDAR_RandLA-Net}	& \textbf{1.2} &  66.5  &      416    &   - & - / -   &53.9   \\
			DRINet~\cite{LiDAR_DRINet} 	& - &  -  &      62    &   -    &- / 2.1 &67.5   \\

			RPVNet~\cite{LiDAR_RPVNet}
			& 24.8 & 119.5   &      168      &    10 & 6.0 / 2.7 &70.3   \\
			
			GASN~\cite{LiDAR_GASN} 	& 2.2  &  \textbf{12.1} &      59     &   -   & - / \textbf{1.6} &70.7   \\
			
			PVKD~\cite{LiDAR_PVKD} 	& -  &  16.1 &      76      &   20  &  9.8 / 3.0 &71.2   \\
			\hline
			
			NUC-Net(Ours)	
			
			&    49.9 &  19.9  &     \textbf{56}    &       \textbf{2.5}   &  \textbf{2.9} / 2.0 &  \textbf{73.6}   \\

			\hline
		\end{tabular}
	\end{center}
	\caption{Results of model efficiency and accuracy on SemanticKITTI test set. The left side of the slash represents GPU memory usage during training, and the right side represents GPU memory usage during testing.}
	\label{tab:efficiency}
	\vspace{-0.5cm} 
\end{table*}

For sparse convolution, the number of MACs is not only determined by the input size and network architecture but also determined by the kernel size which is related to the active synapses. To calculate the MACs of sparse convolution, we follow the work~\cite{LiDAR_NAS} that estimates the average kernel map over the entire dataset and then we calculate the number of MACs as shown in Table \ref{tab:resolution}.

\subsection{Efficiency and Accuracy Comparison}
Table~\ref{tab:efficiency} shows the efficiency and accuracy comparison with the state-of-the-art methods. Thanks to the proposed more efficient and representative NUC representation, our method achieve state-of-the-art performance for accuracy and efficiency. Our method enables smaller input resolution and large batch size for training compared with the uniform counterpart, which reduces the training time from 10 days to 2.5 days compared with the uniform counterpart Cylinder3D. Our method is 36$\%$ inference faster than PVKD and outperforms PVKD by 2.4 mIoU. As PVKD is a knowledge distillation method, it distills the knowledge from a trained model, which requires additional training time. Our method is 8 $\times$ training faster than PVKD. Besides, we can observe that our method reduces training GPU memory usage by 2.3 times compared with Cylinder3D and our method achieves the lowest training GPU memory usage.

\section{Generalize to Multi-scan LiDAR segmentation}
We evaluate our method on SemanticKITTI multi-scan benchmark. The experimental results are shown in Table~\ref{tab:kitti-multi-scan}. We can observe that our method achieves superior performance, and our method outperforms Cylinder3D's multi-scan results with just one-quarter of the voxel input resolution.

\section{Generalize to LiDAR Panoptic Segmentation}
We extend the NUC-Net to panoptic segmentation to show the generalizability of our method.

Panoptic segmentation aims to unify semantic segmentation and instance segmentation and exploit the merits of detection and segmentation. Semantic segmentation is performed for background classes and instance segmentation is performed for foreground classes. The background categories are termed as stuff and the foreground classes are termed as things. We use Panoptic Quality (PQ), Segmentation Quality (SQ) and Recognition Quality (RQ) as the evaluation metrics following~\cite{Panoptic_metric}. Moreover, we report $\rm PQ^{\dag}$ which uses only SQ as PQ in stuff classes.

We use the proposed NUC-Net for panoptic segmentation. In addition, a semantic head is used to predict the semantic labels for stuff categories, and an instance head is used to predict the center heatmap and the offset to the object center.

The experimental results are shown in Table \ref{tab:panoptic}. `Baseline' denotes using the uniform partition method. We can observe that our method achieves state-of-the-art performance on the SemanticKITTI dataset for PQ and mIoU. Non-uniform partition also generalizes well to panoptic segmentation. Note that our method achieves this performance by $120 \times 360 \times 32$ voxel resolution, while Cylinder3D uses the $480 \times 360 \times 32$ voxel resolution.

\section{Generalize to ScribbleKITTI}
ScribbleKITTI~\cite{LiDAR_Scribble} is the first scribble-annotated LiDAR semantic segmentation dataset. To reduce the performance gap  that arises when using such weak annotations, the authors~\cite{LiDAR_Scribble} furhter propose a pipeline to better utilizes scribbles as annotations. This pipeline, combined with existing LiDAR segmentation methods, enhances the performance of baseline segmentation methods when using scribble annotations. To validate the generalizability of the proposed method on the ScribbleKITTI dataset, we use the proposed method as the baseline, incorporating the scribble-based pipeline. The experimental results are shown in Table~\ref{tab:scribble-kitti}. We can observe that our method achieves superior performance. The scribble annotations use geometry line-scribbles to label the point cloud, covering $8.06\%$ of the total number of points, which changes the density of the annotated point cloud. The proposed NUC method aims to  better adapt to the point cloud density variations caused by distance from the sensor. The use of scribble annotations does not fully leverage the strengths of our algorithm.

\section{Generalize to Robo3D}

Robo3D is the first conprehensive benchmark heading toward probing the robustness of 3D detectors and segmentors. We report the Corruption Error (CE) on SemanticKITTI-C dataset. SemanticKTTI-C is derived from the validation set of SemanticKITTI and is constructed using eight corruption types across three severity levels. The eight corruption types are fog, wet ground, snow, motion blur, beam missing, crosstalk, incomplete echo and cross-sensor.

The performance of our method and existing methods are reported in terms of Corruption Error (CE) on the Robo3D dataset as shown in Table~\ref{tab:robo3D}. CE is the primary metric for evaluating model robustness. Additionally, we also report the results of Cylinder3D with an input voxel resolution of [120,360,180] to show the effectiveness of the proposed non-uniform cylindrical partition. We can observe that our method achieves the state-of-the-art performance on model robustness.

By comparing $Cy3D_{SPC}$ and $Cy3D_{[120,360,32]}$, we can observe that after reducing the radial voxel resolution by four times, there is a significant drop in robustness across all corruption types, with the mCE (higher values correspond to worse robustness) increasing from 103.3 to 115.8. Therefore, the reduction in voxel resolution will have a significant impact on the model's robustness. Experimental results show that our method achieves state-of-the-art performance on model robustness with one-quarter voxel resolution by using a reasonable radial interval. Our method achieves $99.9$ mCE using a much lower voxel resolution, which is significant better than the Cylinder3D counterpart with [120, 360,32] resolution and also outperforms Cylinder3D with [480, 360,32] resolution.

\section{Analysis of Non-uniform Cylinderical Partition}

\subsection{Analysis of Performance}

In this section, we analyze why the proposed method outperforms the uniform partition methods and how point distribution affects performance from the following aspects:

\textbf{Encoding Error} Points with different categories may be divided into the same voxel, which results in the encoding error. As shown in Fig.~\ref{fig:vis}, it usually produces wrong predictions at the object boundaries when voxels consist of points from different classes. As shown in Fig.~\ref{fig:encoding_error_distance}, Fig.~\ref{fig:miou_distance} and Fig.~\ref{fig:vis}, our method achieves better performance by reducing the encoding error in the nearby regions.

Since the majority of LiDAR points are within the range of 0-15m and the encoding errors for distant points are small, the increased encoding errors in distant regions have minimal impact on the overall performance. Additionally, our method could enhance the performance of faraway regions by enlarging the receptive field.

\textbf{Receptive Field} Contextual information is vital for LiDAR semantic segmentation due to the following points: (1) LiDAR point cloud has more sparse points in the faraway regions, which requires a larger receptive field for scene understanding. (2) the computation cost is cubically increasing with the kernel size for 3D convolution. To save computation costs, existing works usually adopt small kernel size~\cite{LiDAR_Cylindrical,LiDAR_NAS}, which limits the contextual information. 

\begin{figure}
	\centering
	\includegraphics[width=8cm]{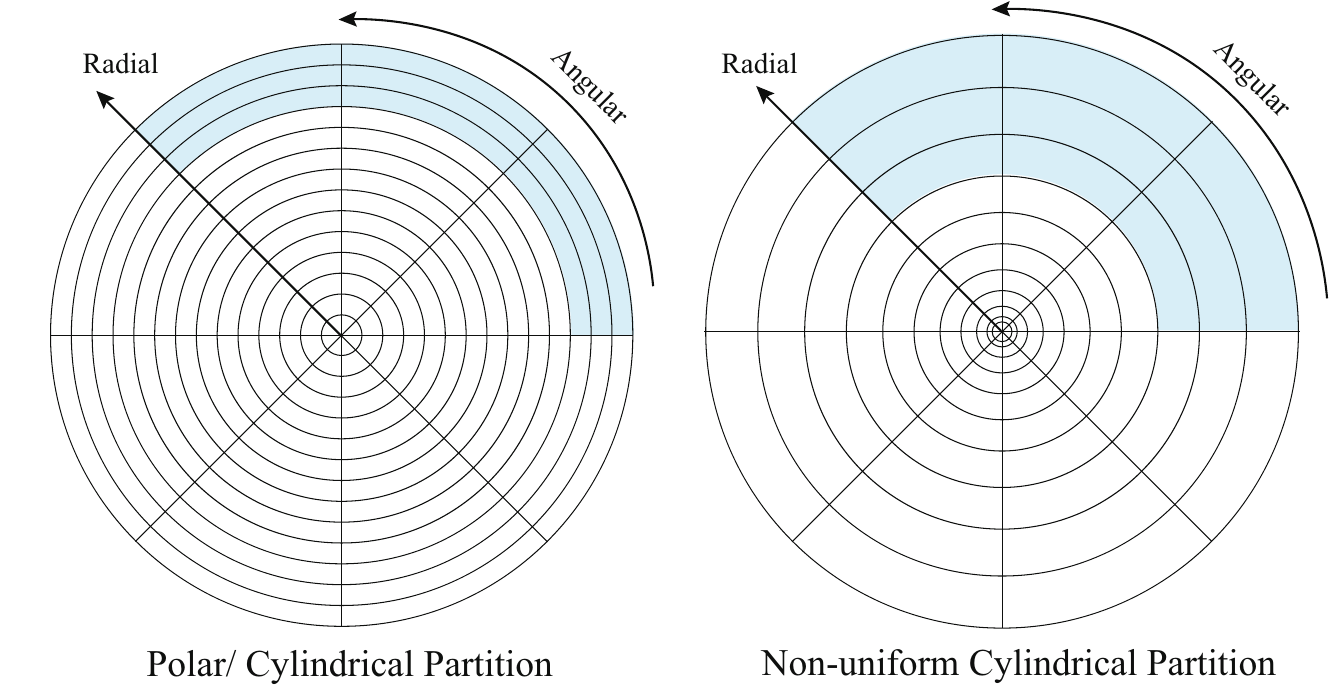} 
	\caption{The receptive field of $3 \times 3$ convolution in different partition methods.}
	\label{fig:receptive_field}
	\vspace{-8pt} 
\end{figure}

\begin{figure}
	\centering
	\includegraphics[width=6cm]{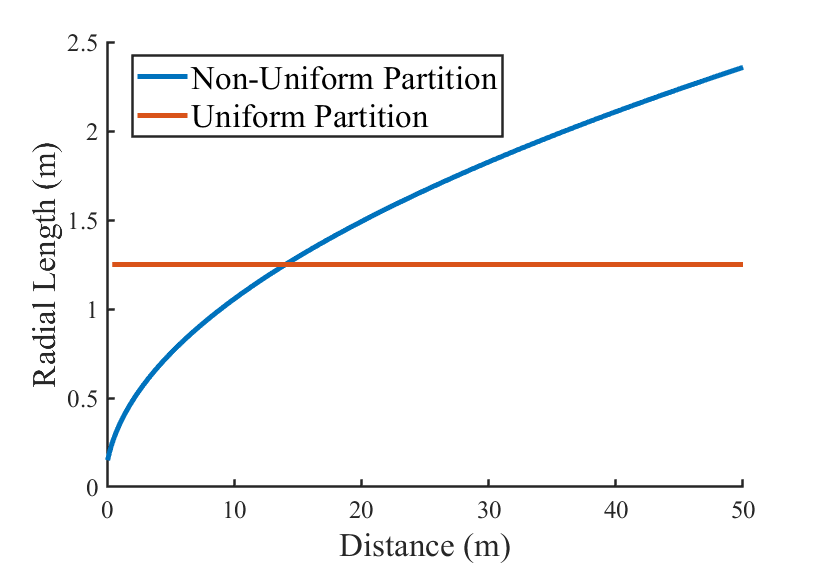} 
	\caption{The variation of radial receptive length of $3 \times 3$ convolution with distance.}
	\label{fig:radial_length}
	\vspace{-0.5cm} 
\end{figure}

Different from the uniform counterpart, our method increases the receptive field without enlarging the size of the convolution kernel, which enhances the performance without adding computation cost. As shown in Fig.~\ref{fig:receptive_field} and Fig.~\ref{fig:radial_length}, we compare the receptive field of a $3 \times 3$ convolution kernel in different partition methods. Our method significantly increases the receptive field in the faraway region compared with the uniform partition methods.

\textbf{Sparse Voxel Representation}
Submanifold Sparse Convolution(SSC) faces the problem of the insufficient receptive field~\cite{sparse_voxel_SST}. SSC does not `fill' the empty voxel, which largely constrains the information communication between voxels. As the proposed non-uniform partition allocates large intervals for the far-away regions, there are fewer empty voxels, which boost the communication between voxels.

\subsection{Analysis of Speedup}

In this section, we analyze the effect of input voxel resolution to network efficiency. Our method applies a mixture of spatially sparse convolution~\cite{sparse_convlution} and submanifold sparse convolution~\cite{submanifold_sparse_convlution} to process the voxel data following~\cite{LiDAR_Cylindrical}. Spatially sparse convolution dilates the sparse data in every layer, while submanifold sparse convolution keeps the same sparsity in every layer. In other words, the use of spatially sparse convolution will yield a rapid growth of the number of active sites (non-empty voxels). For the shallow layer of the network, the speed mainly depends on the number of non-empty voxels, since the non-voxels constitute only a small proportion of overall voxels. For the deep layer of the network, the speed is mainly affected by the voxel resolution.

As shown in Table~\ref{tab:non_empty}, we show the number of non-empty voxels for uniform and non-uniform partition mechanisms. We can observe that the number of the overall non-empty voxels of our method is slightly more than the uniform counterpart and much less than the high-resolution counterpart. The number of non-empty voxels of our method is 1.72 lower than the high-resolution counterpart for the first layer of the 3D network. The use of spatially sparse convolution will yield a rapid growth of the number of active sites (non-empty voxels). With the network deeper, the speedup of NUC is more obvious compared with uniform cylindrical partition.

\section{Conclusion}

In this paper, we propose the non-uniform partition method, named NUC-Net, to process the large-scale LiDAR point cloud. Specifically, we propose the Arithmetic Progression of Interval (API) method to handle the challenges of varied point density of large-scale LiDAR point cloud and propose the non-uniform multi-scale aggregation method to boost the contextual information. Extensive experiments show our method achieves state-of-the-art performance for both accuracy and inference speed. Moreover, our method significantly reduces the training time compared with the uniform counterpart. Our method can be a general component for LiDAR semantic segmentation. We also provide a theoretical analysis of our method. The superior performance is attributed to the NUC-Net which can reduce the encoding error in the nearby regions and enlarge the receptive field in the faraway regions.

\textbf{Limitation and Future Work} Our method shows superior performance on LiDAR point could for both accuracy and efficiency. However, every method has its pros and cons. Cylindrical representation~\cite{LiDAR_Cylindrical,LiDAR_PolarNet} may introduce a bit of distortion compared with cartesian representation, as the voxels in different regions have different occupancies.  In the future, we aim to introduce local geometry structure to merge local geometry information around each point. Since the local geometry is translation invariant and rotation invariant, and free from encoding errors, it can help mitigate the distortion caused by the non-uniform cylindrical partitioning.

\section{Acknowledgement}

This work was supported by the National Key R\&D Program of China under Grant 2023YFF0906200, as well as by the National Natural Science Foundation of China under Grants 62071330 and 62072334.

%



\bibliographystyle{IEEEtran}
\bibliography{IEEEabrv}

\end{document}